\documentclass[preprint,12pt,authoryear]{elsarticle}

\usepackage{amssymb}
\usepackage{makecell}
\usepackage{booktabs}
\usepackage{graphicx}
\usepackage{multirow}
\usepackage{float}
\usepackage{subfig}
\usepackage{color}
\usepackage{amsmath,bm}
\usepackage{changes}
\usepackage{lineno}
\usepackage{array}
\usepackage{algorithm}
\usepackage{algpseudocode}
\usepackage{hyperref}
\usepackage{color, soul}
\setulcolor{red}
\setstcolor{blue}
\sethlcolor{yellow}

\graphicspath{{figures/}}
\newcommand{\up}{$\uparrow$}
\newcommand{\dn}{$\downarrow$}

\begin{document}

\begin{frontmatter}

\title{3D Gaussian Splatting for Fine-grained Urban Scene Reconstruction via Semantic-guided Density Control and Loss Design}

\author[inst1,inst2]{Yueling Liao}
\author[inst2,inst3,inst4]{San Jiang\corref{cor1}}
\author[inst2,inst3,inst4]{Xing Zhang}
\author[inst2,inst3,inst4]{Zhongwen Hu}

\affiliation[inst1]{organization={School of Computer Science, China University of Geosciences (Wuhan)},
            city={Wuhan},
            postcode={430074},
            country={China}}
\affiliation[inst2]{organization={School of Architecture and Urban Planning, Shenzhen University},
            city={Shenzhen},
            postcode={518060},
            country={China}}
\affiliation[inst3]{organization={Guangdong Key Laboratory of Urban Informatics, Shenzhen University},
            city={Shenzhen},
            postcode={518060},
            country={China}}
\affiliation[inst4]{organization={MNR Key Laboratory for Geo-Environmental Monitoring of Great Bay Area, Shenzhen University},
            city={Shenzhen},
            postcode={518060},
            country={China}}

\cortext[cor1]{Corresponding Author: jiangsan@szu.edu.cn}

\begin{abstract}
3D Gaussian Splatting (3DGS) has been becoming a promising 3D reconstruction technique because of its advantage of high-fidelity rendering effect with high efficiency. However, it still suffers from coarse density control and semantically blind loss functions when reconstructing the fine-grained structures in complex urban scenes. Thus, this study proposes SG-GS, a semantic-guided 3DGS method for fine-grained structure reconstruction. It takes the semantic segmentation deduced by the Segment Anything Model (SAM) as a constraint and systematically integrates it into both Gaussian density control and loss function design. On the one hand, a semantic-constrained adaptive density control mechanism is constructed, which dynamically allocates gradient thresholds to different semantic regions of the scene and introduces an additional boundary-enhanced densification strategy, so that new Gaussians are directed to structurally critical regions. On the other hand, a multi-feature guided loss function is designed, and fine-grained supervision signals are gradually introduced through a phased progressive training strategy, alleviating the artifacts caused by the lack of semantic constraints in the basic loss. Experiments on three datasets show that the proposed method consistently improves reconstruction accuracy. Compared with the original 3DGS, the PSNR increases by approximately 2.7\% to 5.2\%, while artifacts on linear structures and boundary blurring are effectively mitigated. The method can serve the fine-grained 3D reconstruction of urban scenes.
\end{abstract}



\end{frontmatter}


\section{Introduction}
\label{sec:intro}

With the advancement of refined urban management, higher requirements are being placed on the accuracy, completeness and richness of detail of 3D urban scene models \citep{jiang2021unmanned,jiang2020efficient,liu2024citygaussian}. Urban scenes are highly complex and contain many semantic categories such as historical buildings, metal towers, vegetation and road markings. The reconstruction targets in different regions differ significantly in geometric features, texture features and spatial structure, which poses a severe challenge to the fine-grained expressive ability of 3D reconstruction methods.

The 3D reconstruction pipeline based on Structure from Motion (SfM) and Multi-View Stereo (MVS) \citep{jiang2024efficient,schonberger2016sfm,furukawa2015mvs,jiang2022parallel} usually executes a sequence of tedious steps, including image orientation, dense point cloud generation, Delaunay meshing and texture mapping. The pipeline is long and computationally heavy, and the resulting mesh models fall short in detail fidelity and novel view synthesis quality. In recent years, Neural Radiance Fields (NeRF) \citep{mildenhall2021nerf} have attracted wide attention for their implicit neural representation and photo-realistic novel view synthesis. However, their computational intensity and long training time \citep{barron2022mipnerf360,turki2022meganerf}, together with an implicit representation that is difficult to edit, limit their large-scale applications.

Currently, 3D Gaussian Splatting (3DGS) \citep{kerbl20233dgs} represents a scene with explicit Gaussian primitives and, combined with highly parallel differentiable rasterization, achieves real-time novel view synthesis while maintaining high rendering quality \citep{yu20263d}. 3DGS has significant advantages over NeRF in rendering efficiency and editability and has become the mainstream method for 3D reconstruction and novel view synthesis \citep{yu2026aerial}. Nevertheless, 3DGS still has two core limitations when facing complex urban structures. First, its Adaptive Density Control (ADC) adopts a globally uniform gradient threshold and cannot apply differentiated densification to thin linear structures (e.g., metal towers and railings) and flat regions (e.g., sky and roads), which leads to under-reconstruction of thin structures and redundant Gaussians in flat regions \citep{ye2024absgs,zhang2024pixelgs}. Second, the training loss imposes the same constraint on all pixels of an image and cannot guide the optimization towards key regions such as building edges and texture details \citep{zhang2024fregs}.

Existing research therefore mainly follows two technical routes, i.e., density control optimization and semantic information fusion. The first category improves representation efficiency by optimizing Gaussian cloning and pruning strategies. AbsGS \citep{ye2024absgs} revealed the gradient collision problem in the original ADC and proposed replacing the vector norm with the sum of absolute gradients. Pixel-GS \citep{zhang2024pixelgs} proposed pixel-area-weighted gradient statistics. 3DGS-MCMC \citep{kheradmand2024mcmc} introduced Markov chain Monte Carlo sampling to replace the traditional ADC. In general, however, these methods adopt globally uniform strategies, ignoring the geometric differences between semantic regions, and can hardly enhance thin structures while compressing flat regions at the same time. The second category focuses on fusing semantic information with 3DGS. Feature 3DGS \citep{zhou2024feature3dgs} attaches extra feature dimensions to each Gaussian and distills 2D features extracted by large models such as SAM/CLIP/DINOv2 into a 3DGS feature field for feature rendering. LangSplat \citep{qin2024langsplat} encodes CLIP language features into Gaussians for open-vocabulary querying. Gaussian Grouping \citep{ye2024gaussiangrouping} groups Gaussians with the instance masks generated by the Segment Anything Model (SAM) \citep{kirillov2023sam} to support object tracking and scene editing. These methods, however, mainly attach semantic attributes to the Gaussians for downstream understanding and editing tasks, which increases the number of parameters and memory footprint without effectively improving the reconstruction accuracy metrics.

To address the insufficient structure preservation, the lack of semantic guidance and the single geometric constraint of existing 3DGS methods in fine-grained urban reconstruction, this paper builds a semantic-guided fine-grained reconstruction framework that jointly optimizes density control and loss supervision. The main contributions are as follows. (1) To address the inability of globally uniform density control to adapt to regional differences in urban scenes, SAM semantic segmentation constraints are introduced into the Gaussian densification process. Gradient thresholds are allocated according to semantic regions and a boundary-enhanced densification mechanism is added, so that newly added Gaussians concentrate on structurally critical regions while redundant generation in flat regions is reduced. (2) To address the inability of a uniform loss function to guide the optimization of key regions, a multi-feature guided loss function is constructed that fuses feature-gradient attention, edge preservation and high-frequency constraints. Combined with a phased progressive training strategy that introduces supervision signals step by step, it improves the fidelity of geometric details while ensuring stable convergence.

This study is organized as follows. Section \ref{sec:related} presents the literature review from density control optimization and semantic information fusion. Section \ref{sec:method} gives the details of the proposed solution for semantic-guided density control and loss design, which is followed by tests in terms of viual analysis of key steps, rendering quality evaluation, and ablation study in Section \ref{sec:exp}. Finally, Section \ref{sec:conclusion} presents conclusions and future studies.

\section{Related work}
\label{sec:related}

\subsection{Density control optimization}
\label{sec:related_density}

In 3DGS, Adaptive Density Control decides where Gaussians are added and removed during optimization \citep{kerbl20233dgs}: the view-space positional gradient of each Gaussian is accumulated over a fixed interval, primitives whose mean gradient exceeds a global threshold are cloned or split, and those with negligible opacity or excessive extent are pruned. This heuristic has well-documented deficiencies. Gradients from different views may cancel out, so that large Gaussians covering complex regions are never densified \citep{ye2024absgs}; the statistics are biased by the number of pixels a Gaussian covers \citep{zhang2024pixelgs}; and because the decision is driven by photometric error alone, flat regions tend to be over-reconstructed while thin structures remain under-reconstructed \citep{bulo2024revising,zhang2024fregs}.

Subsequent work improves ADC along two lines. The first refines the densification criterion. AbsGS \citep{ye2024absgs} replaces the norm of the summed gradient with the sum of absolute gradient components to resolve gradient collision. Pixel-GS \citep{zhang2024pixelgs} weights per-view gradients by pixel coverage and camera distance. \citet{bulo2024revising} reformulate densification from an error-driven perspective and introduce an explicit primitive budget. Taming 3DGS \citep{mallick2024taming} combines gradient, coverage, saliency and depth cues to reach a target budget in a controlled manner. The second line treats density control as a sampling or compression problem. 3DGS-MCMC \citep{kheradmand2024mcmc} replaces cloning and splitting with stochastic Langevin updates and relocation, Mini-Splatting \citep{fang2024minisplatting} and LightGaussian \citep{fan2024lightgaussian} prune primitives according to importance scores to represent a scene with a constrained number of Gaussians, while Scaffold-GS \citep{lu2024scaffoldgs} and CityGaussian \citep{liu2024citygaussian} rely on anchor- or block-based structures to keep the primitive count tractable in large scenes.

Despite their diversity, these methods decide to densify or prune exclusively from image-space statistics, such as gradient magnitude, pixel coverage or photometric error, and apply the same rule uniformly over the scene. In urban environments, however, the required primitive density varies drastically with the semantic category of the surface. Thin towers, railings and cables need many small Gaussians that a global threshold rarely triggers, whereas sky, roads and bare walls are adequately represented by a few large primitives. Globally uniform strategies therefore cannot enhance thin structures and compress flat regions simultaneously, which motivates the semantic-constrained density control proposed in this study.

\subsection{Semantic information fusion}
\label{sec:related_semantic}

Semantic information was first fused into radiance fields in the NeRF literature, where Semantic-NeRF \citep{zhi2021semanticnerf} appends a semantic head to the radiance field and propagates sparse or noisy 2D labels through multi-view consistency. Vision foundation models such as CLIP \citep{radford2021clip}, DINOv2 \citep{oquab2024dinov2} and the Segment Anything Model (SAM) \citep{kirillov2023sam} subsequently made dense 2D semantic priors available for arbitrary scenes, and LERF \citep{kerr2023lerf} distilled language-aligned features into a NeRF for open-vocabulary querying.

The explicit nature of 3DGS makes it convenient to attach such features to individual primitives. Feature 3DGS \citep{zhou2024feature3dgs} augments each Gaussian with a learnable feature vector and distills SAM, CLIP or DINOv2 features into a 3D feature field. LangSplat \citep{qin2024langsplat} uses SAM to define hierarchical semantic levels and compresses CLIP features with a scene-wise autoencoder. LEGaussians \citep{shi2024legaussians} and OpenGaussian \citep{wu2024opengaussian} pursue memory-efficient and point-level open-vocabulary understanding. For segmentation and editing, Gaussian Grouping \citep{ye2024gaussiangrouping} supervises a per-Gaussian identity encoding with video-tracked SAM masks, and SAGA \citep{cen2025saga} distills SAM masks into 3D affinity features. In urban scenes, HUGS \citep{zhou2024hugs} and Street Gaussians \citep{yan2024streetgaussians} jointly model geometry, appearance, semantics and motion to decompose driving sequences into static and dynamic components.

Common to these works is that semantics are treated as an additional \emph{output} of the representation: extra attributes are stored per Gaussian so that semantic maps or language features can be rendered for downstream understanding and editing. This increases the parameter count and memory footprint. A less explored direction uses external priors as additional \emph{supervision}, e.g. monocular depth for geometric regularization \citep{turkulainen2025dnsplatter,yang2024depthanything} or progressive frequency regularization against over-reconstruction \citep{zhang2024fregs}. These priors, however, are semantic-agnostic and constrain every pixel in the same way. To the best of our knowledge, using semantic segmentation priors to decide \emph{where} Gaussians are densified and \emph{how strongly} individual pixels are supervised has not been systematically investigated for 3DGS. The proposed SG-GS (Semantic-Guided 3DGS ) fills this gap by converting SAM region masks into region-wise densification thresholds and a pixel importance weight map, so that semantics guide training without adding any attribute to the final representation.


\begin{figure}[!t]
\centering
\includegraphics[width=\textwidth]{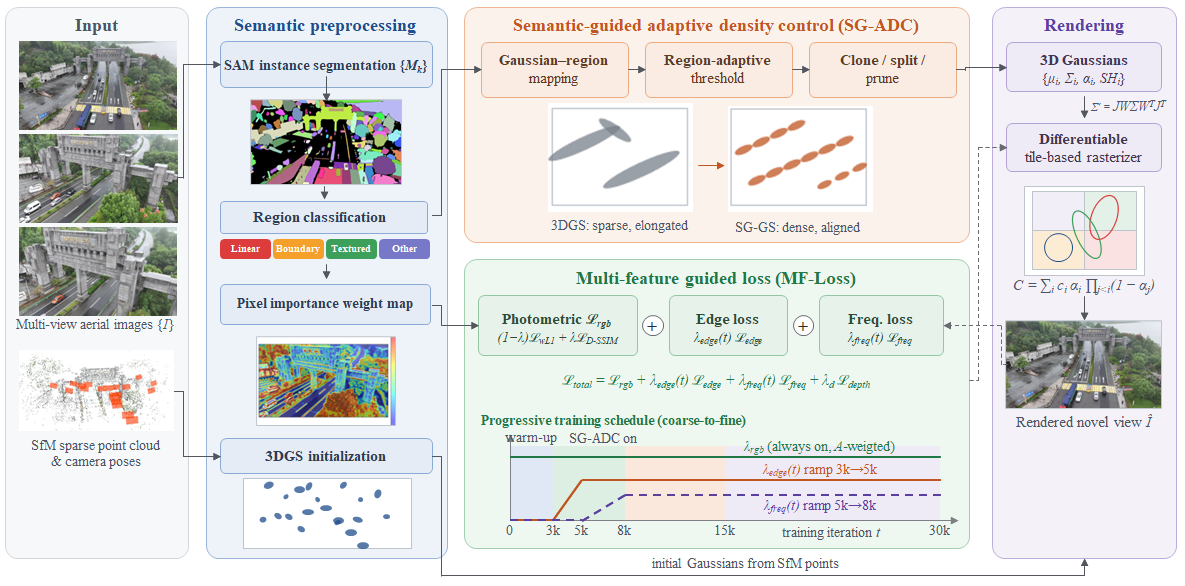}
\caption{Overview of the proposed SG-GS framework. SAM instance masks are classified into four region types and converted into a pixel importance weight map (blue); the region labels drive the semantic-guided adaptive density control (orange) while the weight map drives the multi-feature guided loss with a progressive schedule (green); the optimized Gaussians are rendered by the differentiable rasterizer (purple).}
\label{fig:framework}
\end{figure}

\section{Methodology}
\label{sec:method}

This paper aims to resolve the conflict between insufficient detail representation, accumulation of redundant Gaussians and artifacts in 3DGS reconstruction of urban scenes. Most existing methods optimize from a single perspective such as density control, loss function or regularization, and are prone to sparse reconstruction of linear structures, blurred boundaries and semi-transparent floaters in complex urban scenes. To this end, this study introduces semantic guidance and structural constraints and build the 3DGS framework SG-GS (Semantic-guided Gaussian Splatting), as shown in Fig.~\ref{fig:framework}.

\subsection{3DGS preliminary}
\label{sec:3dgs}

\textbf{Gaussian primitives.} 3DGS is initialized with sparse point clouds reconstructed by SfM and represents the scene explicitly as a set of anisotropic 3D Gaussian primitives. Each primitive is parameterized by a center $\mu$, a 3D covariance matrix $\Sigma$, an opacity $\alpha$ and a set of Spherical Harmonics (SH) coefficients, where the SH coefficients model the view-dependent appearance. A 3D Gaussian primitive with center $\mu$ and covariance $\Sigma$ is defined as
\begin{equation}
G(x)=\exp\!\left(-\tfrac{1}{2}(x-\mu)^{T}\Sigma^{-1}(x-\mu)\right)
\label{eq:gaussian}
\end{equation}
where $x$ is a sample point in 3D space, $\mu$ is the Gaussian center, and $\Sigma$ is the $3\times3$ covariance matrix describing the spatial shape and orientation of the Gaussian primitive.

The covariance matrix must be positive semi-definite to have a clear physical meaning, and directly optimizing its nine elements by gradient descent can hardly satisfy this constraint. Therefore, 3DGS decomposes the covariance into a combination of rotation and scaling:
\begin{equation}
\Sigma = R S S^{T} R^{T}
\label{eq:cov}
\end{equation}
where $R$ is the rotation matrix derived from a unit quaternion and $S$ is the diagonal scaling matrix composed of three axis scaling factors. This parameterization naturally guarantees the positive semi-definiteness of the covariance during optimization, and the rotation and scaling can be optimized independently, so that the Gaussian primitives can adaptively stretch and rotate to fit geometric structures of different scales and orientations compactly.

\textbf{Gaussian rendering.} 3DGS projects the 3D Gaussians onto the 2D image plane. Given the viewing transformation matrix $W$, a local affine approximation of the perspective projection is applied, and the projected 2D covariance is
\begin{equation}
\Sigma' = J W \Sigma W^{T} J^{T}
\label{eq:proj}
\end{equation}
where $J$ is the Jacobian of the affine approximation of the projective transformation. On this basis, 3DGS adopts tile-based differentiable rasterization: the image is divided into $16\times16$-pixel tiles, the Gaussians intersecting each tile are sorted by view-space depth, and forward $\alpha$-blending is then performed along the depth direction for every pixel. The accumulated color is
\begin{equation}
C=\sum_{i\in N} c_{i}\alpha_{i}\prod_{j=1}^{i-1}(1-\alpha_{j})
\label{eq:blend}
\end{equation}
where $N$ is the depth-sorted set of Gaussians covering the pixel, $c_i$ is the view-dependent color of the $i$-th Gaussian computed from its SH coefficients, and $\alpha_i$ is obtained by multiplying the opacity of the Gaussian with the density of its projected 2D Gaussian at that pixel. The whole rasterization process is highly parallel and fully differentiable, which is the key for 3DGS to achieve real-time high-quality rendering.

\textbf{Model training.} 3DGS is supervised by the photometric error between the rendered image and the ground-truth image. The loss is a fixed weighted combination of an $\mathcal{L}_1$ term and a structural similarity term $\mathcal{L}_{\text{D-SSIM}}$:
\begin{equation}
\mathcal{L}=(1-\lambda)\mathcal{L}_{1}+\lambda\mathcal{L}_{\text{D-SSIM}}
\label{eq:loss3dgs}
\end{equation}
where $\lambda$ is the balancing coefficient, set to 0.2 in the official implementation. Since the initial SfM point cloud is sparse and unevenly distributed, parameter optimization alone can hardly cover under-reconstructed regions. 3DGS therefore periodically performs adaptive density control during training. For Gaussians whose accumulated mean view-space positional gradient exceeds a threshold, cloning (for small Gaussians, filling under-reconstructed regions) or splitting (for large Gaussians, refining over-reconstructed regions) is executed according to their scale; opacities are reset periodically, and Gaussians with low opacity or excessive size are pruned to suppress redundancy and floaters. The above density control criterion and loss design constitute the baseline framework improved in this study.

\subsection{Semantic-constrained adaptive density control}
\label{sec:sgadc}

The original 3DGS adopts a globally uniform density control strategy: the mean view-space positional gradient of each Gaussian primitive is accumulated during training, and densification is triggered when it exceeds a fixed global threshold $\tau$. Let $s_{\max,i}$ be the maximum scale of the $i$-th Gaussian, $\rho$ the percent-dense parameter and $d$ the spatial extent of the scene; the original densification criterion is
\begin{equation}
\mathrm{densify}(g_{i})=
\begin{cases}
\mathrm{clone}, & \left\|\nabla_{\mu'}L_{i}\right\|\ge\tau\ \text{and}\ s_{\max,i}\le\rho\cdot d\\[2pt]
\mathrm{split}, & \left\|\nabla_{\mu'}L_{i}\right\|\ge\tau\ \text{and}\ s_{\max,i}>\rho\cdot d
\end{cases}
\label{eq:densify}
\end{equation}
where $\nabla_{\mu'}L_i$ denotes the accumulated mean loss gradient of Gaussian $g_i$ with respect to its projected view-space position. This criterion treats all Gaussians in the scene equally and cannot adapt to the pronounced regional differences of urban scenes.

To alleviate this issue, we introduce semantic prior on top of gradient-driven density control and construct a semantic-aware adaptive density control mechanism. This section details the Semantic-Guided Adaptive Density Control (SG-ADC), whose workflow consists of semantic region classification, Gaussian-region mapping and adaptive-threshold density control.

\subsubsection{Semantic region classification}

In the preprocessing stage, the automatic mask generator of SAM is first run on all training images to obtain a set of instance-level binary masks $\{M_k\}$. Owing to its strong zero-shot generalization, SAM can accurately delineate the instance boundaries of various objects in urban scenes without any scene-specific annotation. However, the instance masks themselves carry no information about reconstruction difficulty. We therefore classify the masks, according to their shape and texture statistics, into four difficulty regions: linear structures (e.g., towers), boundary regions (instance-to-instance boundaries), textured surfaces (e.g., trees) and other regions. The classification relies on three statistics: compactness, aspect ratio and gradient magnitude.

\textbf{Compactness} measures how far the mask shape deviates from a circle; linear structures have markedly low compactness:
\begin{equation}
C_{k}=\frac{4\pi A_{k}}{P_{k}^{2}+\varepsilon}
\label{eq:compact}
\end{equation}
where $A_k$ is the pixel area of mask $M_k$, $P_k$ its contour perimeter and $\varepsilon$ a numerical stability term.

\textbf{Aspect ratio} measures the elongation of the mask:
\begin{equation}
R_{k}=\frac{\max(w_{k},h_{k})}{\min(w_{k},h_{k})}
\label{eq:aspect}
\end{equation}
where $w_k$ and $h_k$ are the side lengths of the minimum-area rotated bounding rectangle of the mask.

\textbf{Mean gradient magnitude of a region} measures the richness of texture inside the mask:
\begin{equation}
G_{k}=\frac{1}{|M_{k}|}\sum_{(x,y)\in M_{k}}\sqrt{G_{x}^{2}(x,y)+G_{y}^{2}(x,y)}
\label{eq:gradmag}
\end{equation}
where $G_x$ and $G_y$ are the responses of the edge operator in the horizontal and vertical directions, respectively.

\begin{figure}[!t]
\centering
\includegraphics[width=\textwidth]{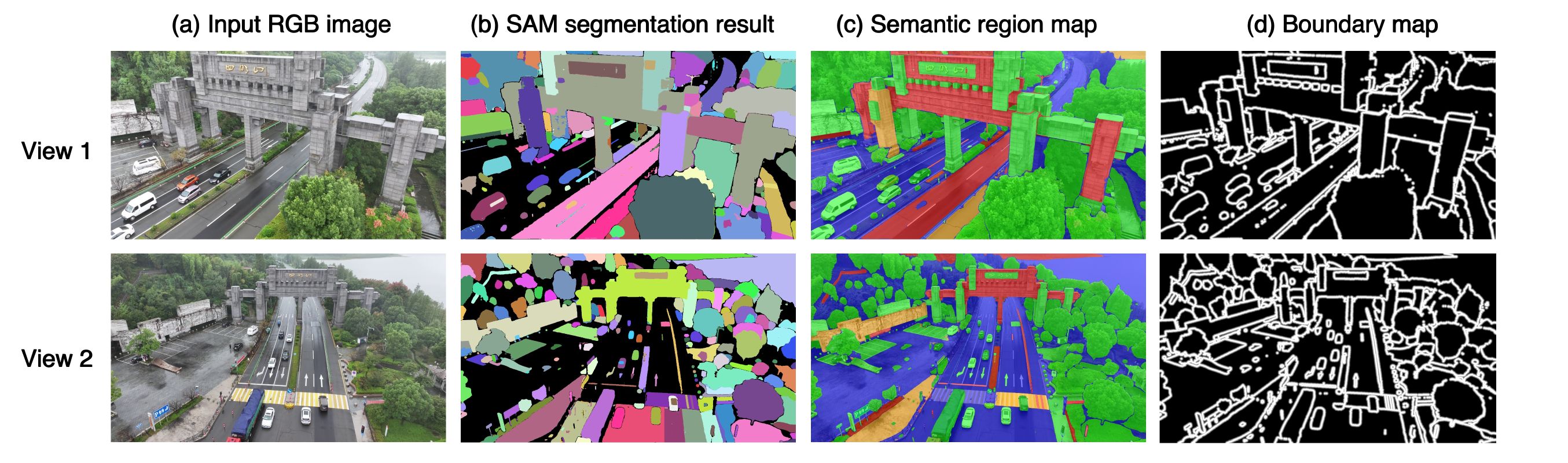}
\caption{Visualization of SAM semantic region classification.}
\label{fig:samvis}
\end{figure}

Based on the three statistics, the region classification rule is defined as
\begin{equation}
l_{k}=
\begin{cases}
\text{linear structure}, & C_{k}<\theta_{C}^{(1)}\ \text{or}\ R_{k}>\theta_{R}^{(1)}\\[2pt]
\text{boundary region}, & C_{k}<\theta_{C}^{(2)}\ \text{and}\ R_{k}>\theta_{R}^{(2)}\\[2pt]
\text{textured surface}, & G_{k}>\theta_{g}\\[2pt]
\text{other region}, & \text{otherwise}
\end{cases}
\label{eq:classify}
\end{equation}
where $\theta_C^{(1)}$, $\theta_C^{(2)}$, $\theta_R^{(1)}$, $\theta_R^{(2)}$ and $\theta_g$ are the classification thresholds for compactness, aspect ratio and gradient magnitude. The rules are evaluated in order: linear and boundary regions are first identified by shape features, and textured surfaces are then separated from other regions by texture strength. The classified masks are painted onto a pixel-level region map in descending order of area, so that small structures are not swallowed by large backgrounds; pixels not covered by any mask default to the other region. Fig.~\ref{fig:samvis} visualizes the SAM segmentation and region classification process, including the SAM segmentation result, the semantic region map and the boundary map.

\subsubsection{Gaussian--region mapping}

\begin{figure}[!t]
\centering
\includegraphics[width=0.7\textwidth]{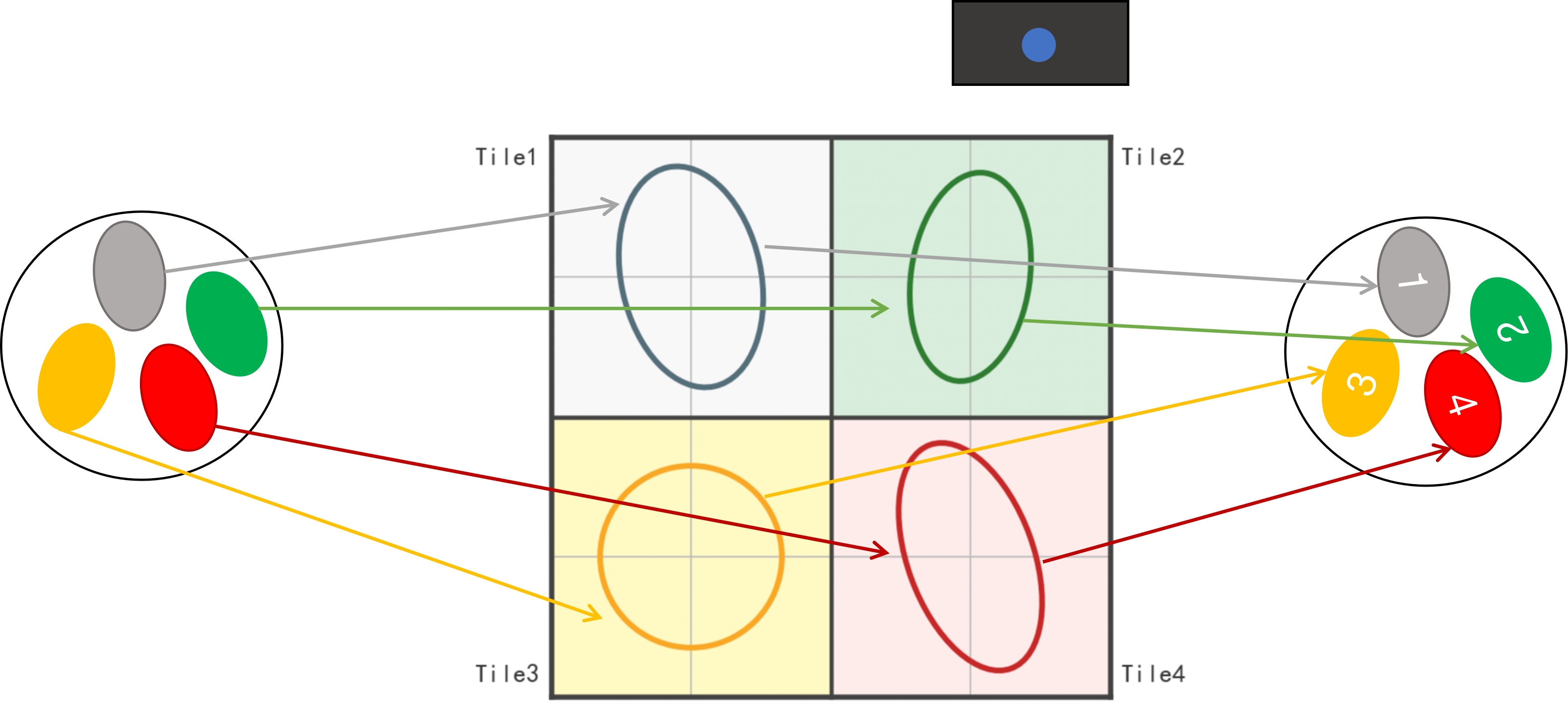}
\caption{Illustration of the Gaussian--region mapping.}
\label{fig:mapping}
\end{figure}

The region map and the boundary proximity map are defined in the 2D image domain, whereas density control operates on Gaussian primitives in 3D space. A mapping between the two must therefore be established. As shown in Fig.~\ref{fig:mapping}, for the camera of the current view during training, each Gaussian center $\mu_i\in\mathbb{R}^3$ is projected to pixel coordinates through the full projection matrix $P_{\text{proj}}$ (world to clip space):
\begin{equation}
\begin{aligned}
[x'_{i},y'_{i},z'_{i},w_{i}]^{\top}&=P_{\text{proj}}[\mu_{i},1]^{\top},
u_{i}=\frac{x'_{i}/w_{i}+1}{2}\,W,
v_{i}=\frac{y'_{i}/w_{i}+1}{2}\,H
\end{aligned}
\label{eq:projection}
\end{equation}
where $W$ and $H$ are the image width and height. A projection is valid if $z'_i>0$ (the Gaussian lies in front of the camera) and $(u_i,v_i)$ falls inside the image. For validly projected Gaussians, the region type $r_i$ and boundary proximity $b_i$ are obtained by nearest-neighbor lookup, i.e., $r_i=S(\lfloor v_i\rfloor,\lfloor u_i\rfloor)$ and $b_i=B(\lfloor v_i\rfloor,\lfloor u_i\rfloor)$. Gaussians with invalid projections (outside the frustum or behind the camera) are assigned the textured-surface class by default, i.e., the standard threshold, so that their densification behavior is identical to the original 3DGS and the method never degrades below the baseline.

\subsubsection{Adaptive-threshold density control}

\begin{figure}[!t]
\centering
\includegraphics[width=0.85\textwidth]{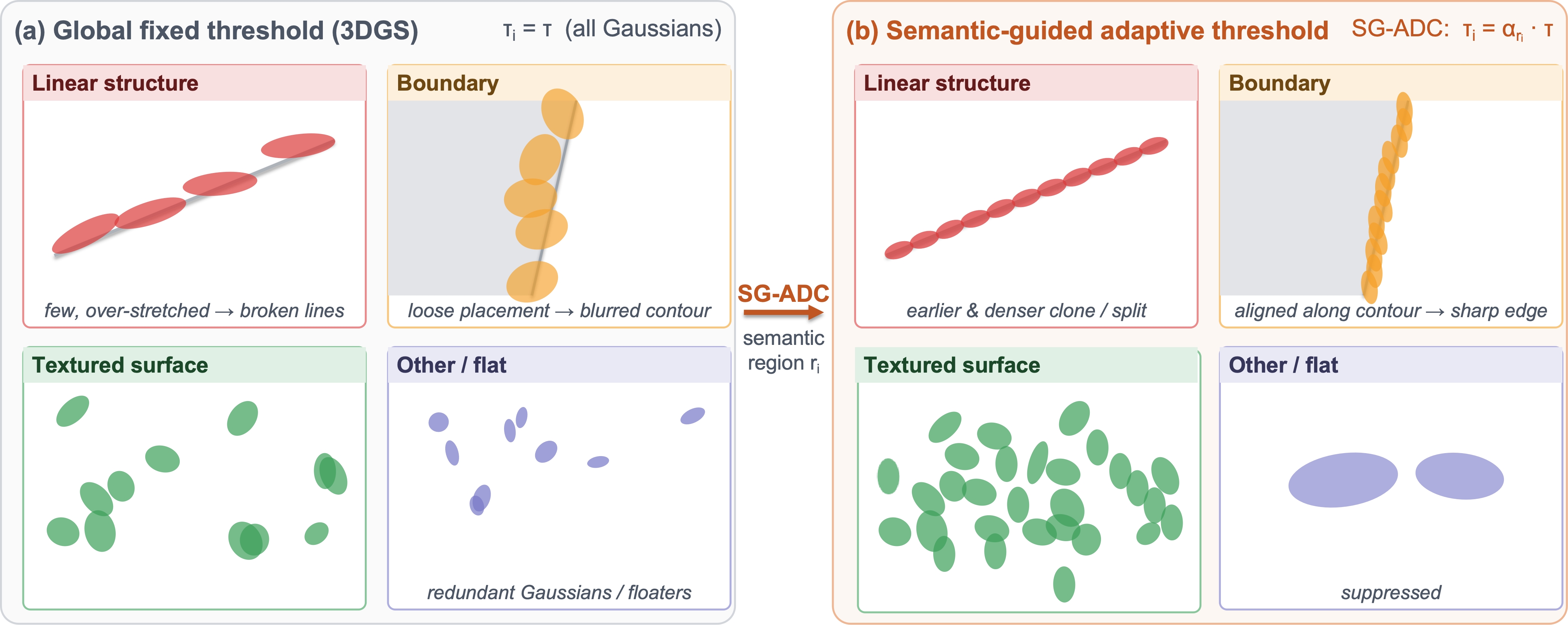}
\caption{Adaptive-threshold density control strategy.}
\label{fig:adc}
\end{figure}

Building on the Gaussian--region mapping, the global threshold $\tau$ in Eq.~(\ref{eq:densify}) is replaced by a per-Gaussian threshold scaled according to the region type, as given in Eq.~(\ref{eq:tau}), where $\alpha_{r_i}$ is the threshold scaling factor of region $r_i$ and satisfies $\alpha_{\text{line}}<\alpha_{\text{edge}}<\alpha_{\text{tex}}<\alpha_{\text{other}}$. Lowering the threshold in linear-structure regions makes thin structures with weak gradients trigger cloning and splitting earlier and more frequently, alleviating under-reconstruction; the threshold in boundary regions is moderately lowered to sharpen contours; textured surfaces keep the standard threshold; and the threshold in other regions is raised to suppress redundant Gaussians in flat areas such as road surfaces.
\begin{equation}
\tau_{i}=\alpha_{r_{i}}\cdot\tau
\label{eq:tau}
\end{equation}
Accordingly, the densification criterion changes from a uniform test to the per-Gaussian test $\|\nabla_{\mu'}L_i\|\ge\tau_i$, while the size branch conditions for cloning and splitting remain unchanged, thereby reallocating the budget of new Gaussians to structurally critical regions. Besides, since the Gaussian distribution has not converged in early training and the projected positions jitter considerably, region lookups are noisy at that stage. We therefore keep the original density control in the early phase and enable the semantic-guided adaptive threshold and boundary enhancement only after the geometric skeleton has stabilized, so that early classification noise does not mislead the densification. The proposed density control strategy is illustrated in Fig.~\ref{fig:adc}.

\subsection{Multi-feature guided loss and progressive training}
\label{sec:mfloss}

The original 3DGS uses a fixed weighted combination of L1 and D-SSIM as the reconstruction loss, applying supervision of equal strength to all pixels, which is semantically blind: low-information regions such as sky and road surfaces usually occupy most pixels of urban images, so the optimization capacity is spread over these regions, while key details that carry visual quality, such as building contours, receive insufficient supervision, resulting in blurred edges and lost high-frequency textures.

We design the Multi-Feature Guided Loss (MF-Loss). In the preprocessing stage, a pixel importance weight map is built offline from the multi-granularity features of SAM; during training, the reconstruction loss is weighted per pixel accordingly, an edge-preservation constraint and a frequency-domain high-frequency constraint are added, and a phased progressive training strategy controls when and how strongly each supervision signal is introduced. Unlike methods that attach learnable semantic features to the Gaussian primitives, this design places all guidance on the loss side, adds no model parameters or rendering overhead, and is thus a lightweight external guidance.

\subsubsection{Pixel importance weight map}

The key idea of the weight map is that the reconstruction difficulty differs significantly between semantic regions: linear structures and semantic boundaries are where errors concentrate and supervision should be biased, whereas flat regions such as sky and roads only need basic constraints.

First, a base importance is assigned to every pixel according to its region type:
\begin{equation}
W_{\text{sem}}(x,y)=\omega_{S(x,y)}
\label{eq:wsem}
\end{equation}
where $S(x,y)$ is the semantic region type of pixel $(x,y)$ and $\omega_r\in(0,1]$ is the importance coefficient of region $r$.

Second, the base weight is enhanced with the boundary proximity map $B$ and bounded from below and above, yielding the final pixel importance weight map:
\begin{equation}
A(x,y)=a_{\min}+(1-a_{\min})\min\!\left(W_{\text{sem}}(x,y)+\beta B(x,y),\,1\right)
\label{eq:weightmap}
\end{equation}
where $\beta$ is the boundary gain and $a_{\min}$ is the lower bound of the weight. The lower bound guarantees that other regions still receive basic supervision and prevents training degeneration caused by low-weight regions losing all constraints.

\subsubsection{Multi-feature guided weighted loss}

MF-Loss consists of four parts: a weighted L1 term, a structural similarity term, an edge-preservation term and a frequency-domain high-frequency term. The per-pixel absolute error is weighted by the weight map $A$ so that the optimization capacity is biased towards high-importance regions:
\begin{equation}
\mathcal{L}_{wL1}=\frac{1}{3HW}\sum_{c=1}^{3}\sum_{x,y}A(x,y)\left|\hat{I}_{c}(x,y)-I_{c}(x,y)\right|
\label{eq:wl1}
\end{equation}
where $\hat{I}$ and $I$ are the rendered and ground-truth images and $c$ indexes the color channels. Since SSIM itself measures local structural similarity, it is not weighted per pixel; it is combined with the weighted L1 term to form the RGB main loss:
\begin{equation}
\mathcal{L}_{\text{rgb}}=(1-\lambda)\mathcal{L}_{wL1}+\lambda\left(1-\mathrm{SSIM}(\hat{I},I)\right)
\label{eq:lrgb}
\end{equation}
where $\lambda$ is the balancing coefficient of the structural similarity term. To constrain the consistency of edge structures, the images are first converted to grayscale and the edge response magnitude $E(\cdot)$ is extracted with an edge operator; the edge-preservation loss is defined as the L1 distance between the edge maps of the rendered and ground-truth images:
\begin{equation}
\mathcal{L}_{\text{edge}}=\left\|E(\hat{I})-E(I)\right\|_{1}
\label{eq:ledge}
\end{equation}
This term directly penalizes deviations in edge position and strength, suppressing blur and misalignment at building contours and linear structures. High-frequency details are easily averaged out under spatial-domain losses; therefore, a frequency-domain constraint is applied only within the high-importance region $\mathcal{H}=\{(x,y):A(x,y)>\theta_A\}$. The masked grayscale images are transformed by the 2D Fourier transform and their log-magnitude spectra are compared:
\begin{equation}
\mathcal{L}_{\text{freq}}=\left\|\log\!\left(\left|\mathcal{F}(\hat{I}\odot\mathbf{1}_{\mathcal{H}})\right|+\epsilon\right)-\log\!\left(\left|\mathcal{F}(I\odot\mathbf{1}_{\mathcal{H}})\right|+\epsilon\right)\right\|_{1}
\label{eq:lfreq}
\end{equation}
where $\theta_A$ is the importance threshold, $\mathcal{F}$ is the 2D fast Fourier transform, $\odot$ denotes element-wise multiplication and $\epsilon$ is a numerical stability term. The log-magnitude spectrum compresses the dynamic range of low-frequency components, making the loss more sensitive to differences in high-frequency components. Combining the above terms, the total loss is defined as
\begin{equation}
\mathcal{L}_{\text{total}}=\mathcal{L}_{\text{rgb}}+\lambda_{\text{edge}}(t)\mathcal{L}_{\text{edge}}+\lambda_{\text{freq}}(t)\mathcal{L}_{\text{freq}}+\lambda_{d}\mathcal{L}_{\text{depth}}
\label{eq:ltotal}
\end{equation}
where $\mathcal{L}_{\text{depth}}$ follows the depth regularization term of the official 3DGS implementation with depth priors provided by Depth Anything V2 \citep{yang2024depthanything}, $\lambda_d$ is its weight and $t$ is the iteration index.

\subsubsection{Phased progressive training strategy}

\begin{figure}[!t]
\centering
\includegraphics[width=0.9\textwidth]{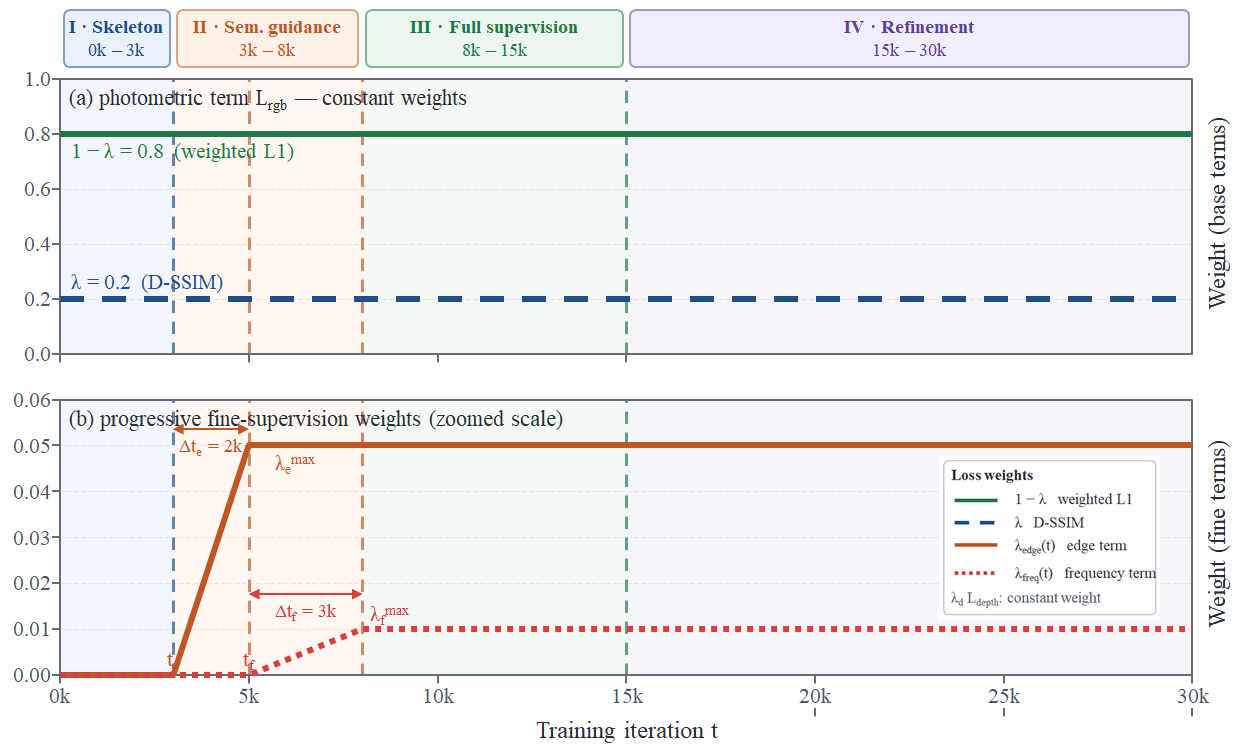}
\caption{Weight schedule curves of the phased progressive training of MF-Loss.}
\label{fig:schedule}
\end{figure}

Rendered results are noisy in early training. If fine-grained supervision such as edge and frequency losses is introduced too early, their gradients are dominated by the early rendering noise, which amplifies errors and destabilizes convergence. The weights of the edge and frequency losses therefore follow a linear ramp schedule that introduces fine supervision progressively:
\begin{equation}
\begin{split}
\lambda_{\text{edge}}(t)=\lambda_{e}^{\max}\cdot\left[\min\!\left(1,\frac{t-t_{e}}{\Delta t_{e}}\right)\right]_{+} \\ 
\lambda_{\text{freq}}(t)=\lambda_{f}^{\max}\cdot\left[\min\!\left(1,\frac{t-t_{f}}{\Delta t_{f}}\right)\right]_{+}
\label{eq:schedule}
\end{split}
\end{equation}
where $\lambda_e^{\max}$ and $\lambda_f^{\max}$ are the upper bounds of the edge and frequency loss weights, $t_e$ and $t_f$ the corresponding starting iterations, $\Delta t_e$ and $\Delta t_f$ the lengths of the ramp intervals, and $[\cdot]_+$ truncates negative values to zero. The schedule curves are shown in Fig.~\ref{fig:schedule}.

\begin{algorithm}[!t]
\small
\caption{\textbf{Semantic-guided loss $\mathcal{L}_{\text{total}}$ and progressive training}}
\label{alg:mfloss}
\begin{algorithmic}[1]
\Statex \textbf{Input:} \begin{minipage}[t]{\dimexpr\linewidth-5em}\raggedright
Training images $\{I_t\}_{t=1}^{T}$; semantic region maps $\{S_t\}$ and boundary proximity maps $\{B_t\}$;
region importance coefficients $\{\omega_r\}$, boundary gain $\beta$, weight lower bound $a_{\min}$;
SSIM balancing coefficient $\lambda$, high-importance threshold $\theta_A$, depth regularization weight $\lambda_d$;
ramp parameters $(\lambda_e^{\max},t_e,\Delta t_e)$ and $(\lambda_f^{\max},t_f,\Delta t_f)$; total number of iterations $N_{\text{iter}}$
\end{minipage}
\Statex \textbf{Output:} Optimized set of Gaussian primitives $\mathcal{G}$
\Statex \textbf{Stage 1: Preprocessing --- pixel importance weight-map construction}
\For{$t = 1, 2, \ldots, T$}
\State $A_t \leftarrow a_{\min}+(1-a_{\min})\min\!\big(\omega_{S_t}+\beta\, B_t,\,1\big)$
\Comment{boundary-enhanced weights}
\State $\textsc{SaveHalf}(A_t)$
\Comment{offline half-precision storage}
\EndFor
\Statex \textbf{Stage 2: Training --- weighted loss and progressive scheduling}
\For{$n = 1, 2, \ldots, N_{\text{iter}}$}
\State Sample view $t$, render $\hat{I}$, load ground truth $I \leftarrow I_t$ and weight map $A \leftarrow A_t$
\State $\mathcal{L}_{wL1} \leftarrow \tfrac{1}{3HW}\sum_{c}\sum_{x,y}A(x,y)\,\lvert\hat{I}_c-I_c\rvert$
\Comment{weight-map weighted L1}
\State $\mathcal{L}_{\text{rgb}} \leftarrow (1-\lambda)\,\mathcal{L}_{wL1}+\lambda\big(1-\mathrm{SSIM}(\hat{I},I)\big)$
\Comment{RGB main loss}
\State $\lambda_{\text{edge}} \leftarrow \lambda_e^{\max}\big[\min(1,(n-t_e)/\Delta t_e)\big]_{+}$
\Comment{edge-weight ramp}
\State $\lambda_{\text{freq}} \leftarrow \lambda_f^{\max}\big[\min(1,(n-t_f)/\Delta t_f)\big]_{+}$
\Comment{frequency-weight ramp}
\If{$\lambda_{\text{edge}}>0$}
\State $\mathcal{L}_{\text{edge}} \leftarrow \big\lVert E(\hat{I})-E(I)\big\rVert_1$
\Comment{edge-preservation loss}
\EndIf
\If{$\lambda_{\text{freq}}>0$}
\State $\mathcal{H} \leftarrow \{(x,y):A(x,y)>\theta_A\}$
\Comment{high-importance region mask}
\State $\mathcal{L}_{\text{freq}} \leftarrow \big\lVert\log(|\mathcal{F}(\hat{I}\odot\mathbf{1}_{\mathcal{H}})|+\epsilon)-\log(|\mathcal{F}(I\odot\mathbf{1}_{\mathcal{H}})|+\epsilon)\big\rVert_1$
\Comment{frequency loss}
\EndIf
\State $\mathcal{L}_{\text{total}} \leftarrow \mathcal{L}_{\text{rgb}}+\lambda_{\text{edge}}\,\mathcal{L}_{\text{edge}}+\lambda_{\text{freq}}\,\mathcal{L}_{\text{freq}}+\lambda_d\,\mathcal{L}_{\text{depth}}$
\Comment{total loss}
\State $\mathcal{G} \leftarrow \textsc{Update}(\mathcal{G})$
\EndFor
\State \Return $\mathcal{G}$
\end{algorithmic}
\end{algorithm}

Together with the semantic-guided density control, the whole training process evolves progressively from coarse to fine. In the early phase, only the weight-map RGB loss and the depth regularizer are used together with the original density control to establish a stable geometric skeleton; SG-ADC and boundary enhancement are then enabled and the edge loss ramps up linearly; once the geometry has essentially converged, the frequency-domain high-frequency constraint is introduced; finally, densification is turned off and all loss weights remain at their full values until convergence. This strategy cooperates with the ``skeleton first, details later'' rhythm of SG-ADC and improves detail fidelity step by step while keeping convergence stable.

In summary, the complete computation flow of Section~\ref{sec:mfloss} is given in Algorithm~\ref{alg:mfloss}. The pixel important weight maps are constructed offline using the SAM segmentation priors during the preprocessing phase. Each iteration in the training phase reads the current view's weight map, calculates the weighted RGB main loss, adds the frequency-domain high-frequency and edge-preservation requirements in accordance with the linear ramp schedule, then back-propagates the total loss to update the Gaussian parameters.

\section{Experiments and Analysis}
\label{sec:exp}

\subsection{Datasets and configuration}
\label{sec:datasets}

\begin{figure}[!t]
\centering
\includegraphics[width=0.8\textwidth]{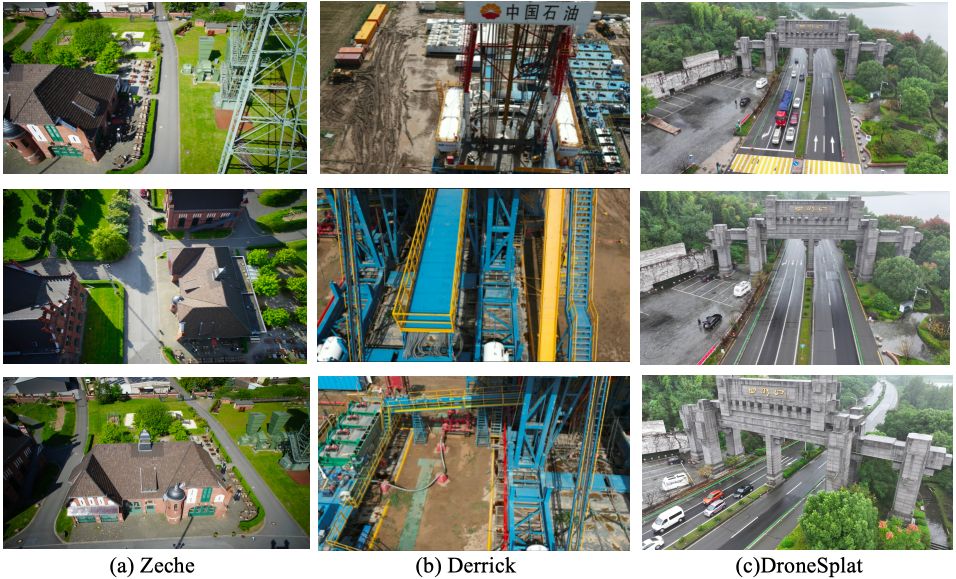}
\caption{Example images of the datasets.}
\label{fig:datasets}
\end{figure}

To verify the effectiveness of the proposed method, experiments are conducted on the public Zeche \citep{nex2015isprs} and DroneSplat \citep{tang2025dronesplat} datasets and on the self-collected Derrick dataset. Details of the datasets are listed in Table~\ref{tab:datasets}. The Zeche dataset contains 147 aerial images with many object instances and complex texture structures, and is used to verify the detail reconstruction ability in complex urban scenes. The Derrick dataset contains 292 aerial images with a large number of steel linear structures, and is used to verify the reconstruction of thin geometric structures. The DroneSplat dataset contains 24 sequences captured by drones, covering dynamic and static scenes, and is used to verify static scene reconstruction from drone imagery in the presence of dynamic distractors. For all datasets, COLMAP is used for feature extraction, incremental SfM reconstruction and camera pose estimation, and the resulting sparse point clouds serve as the initialization of 3DGS. Example images of the datasets are shown in Fig.~\ref{fig:datasets}.

\begin{table}[!t]
\centering
\caption{Details of the datasets.}
\label{tab:datasets}
\footnotesize
\setlength{\tabcolsep}{4pt}
\begin{tabular}{l>{\raggedright\arraybackslash}p{2.5cm}>{\raggedright\arraybackslash}p{2.5cm}>{\raggedright\arraybackslash}p{2.3cm}}
\toprule
\textbf{Name} & \textbf{Zeche} & \textbf{Derrick} & \textbf{DroneSplat}\\
\midrule
Number of images & 147 & 292 & 24\\
POS data & Yes & Yes & Yes\\
Number of sparse points ($\times10^{4}$) & 15 & 20.8 & 1.6\\
Image size (pixels) & 8,300$\times$5,500 & 6,000$\times$4,000 & 1,920$\times$1,080\\
Train/test split & 128/19 & 255/37 & ---\\
\bottomrule
\end{tabular}
\end{table}

\begin{table}[!t]
\centering
\caption{Parameter settings.}
\label{tab:params}
\begin{tabular}{ll}
\toprule
\textbf{Parameter} & \textbf{Value}\\
\midrule
Rendering resolution (-r) & 1\\
Training iterations & 30,000\\
SH degree & 3\\
Base gradient threshold $\tau$ & 0.0002\\
Region threshold scaling factors $\alpha$ & 0.5 / 0.7 / 0.9 / 1.5\\
Gradient magnitude threshold $\theta_g$ & 30\\
SG-ADC start iteration & 3,000\\
SSIM balancing coefficient $\lambda$ & 0.2\\
Densification stop iteration & 15,000\\
\bottomrule
\end{tabular}
\end{table}

\begin{table}[!t]
\centering
\caption{Evaluation metrics.}
\label{tab:metrics}
\small
\begin{tabular}{>{\raggedright\arraybackslash}p{5.6cm}>{\raggedright\arraybackslash}p{6.6cm}}
\toprule
\textbf{Metric} & \textbf{Description}\\
\midrule
Structural Similarity Index (SSIM) & Similarity measure based on image structural information\\
Peak Signal-to-Noise Ratio (PSNR) & Objective criterion based on the mean squared error\\
Learned Perceptual Image Patch Similarity (LPIPS) & Image similarity measure based on human perception\\
\bottomrule
\end{tabular}
\end{table}

The experimental server runs Ubuntu 22.04 with a single NVIDIA GeForce RTX 4090 GPU; the deep learning framework is PyTorch 2.1.2 with CUDA 11.8. The model is trained for 30,000 iterations. The original density control is used for the first 3,000 iterations to establish a stable geometric skeleton. From iteration 3,000, the semantic-guided adaptive threshold and the boundary-enhanced densification strategy are enabled, with boundary enhancement executed every 500 iterations; meanwhile, the edge-preservation loss ramps up linearly to its full value within 2,000 iterations starting from iteration 3,000. The frequency-domain high-frequency loss ramps up linearly to its full value within 3,000 iterations starting from iteration 5,000. Densification is turned off at iteration 15,000, after which all loss weights remain at their full values and the model enters the convergence refinement phase until the end of training. The semantic region map, the boundary proximity map and the pixel importance weight map are all generated offline in the preprocessing stage. After training, the model is quantized and saved. The parameter settings are listed in Table~\ref{tab:params}.

Three metrics used for 3DGS are adopted for evaluation, i.e., Peak Signal-to-Noise Ratio (PSNR), Structural Similarity (SSIM) and Learned Perceptual Image Patch Similarity (LPIPS). A higher PSNR indicates better image quality, a higher SSIM indicates more similar images, and a lower LPIPS indicates higher perceptual similarity. The metrics are listed in Table~\ref{tab:metrics}.

\subsection{Visual analysis of key individual modules}
\label{sec:keysteps}

\subsubsection{SAM segmentation and region classification}
\begin{figure}[!t]
\centering
\includegraphics[width=0.85\textwidth]{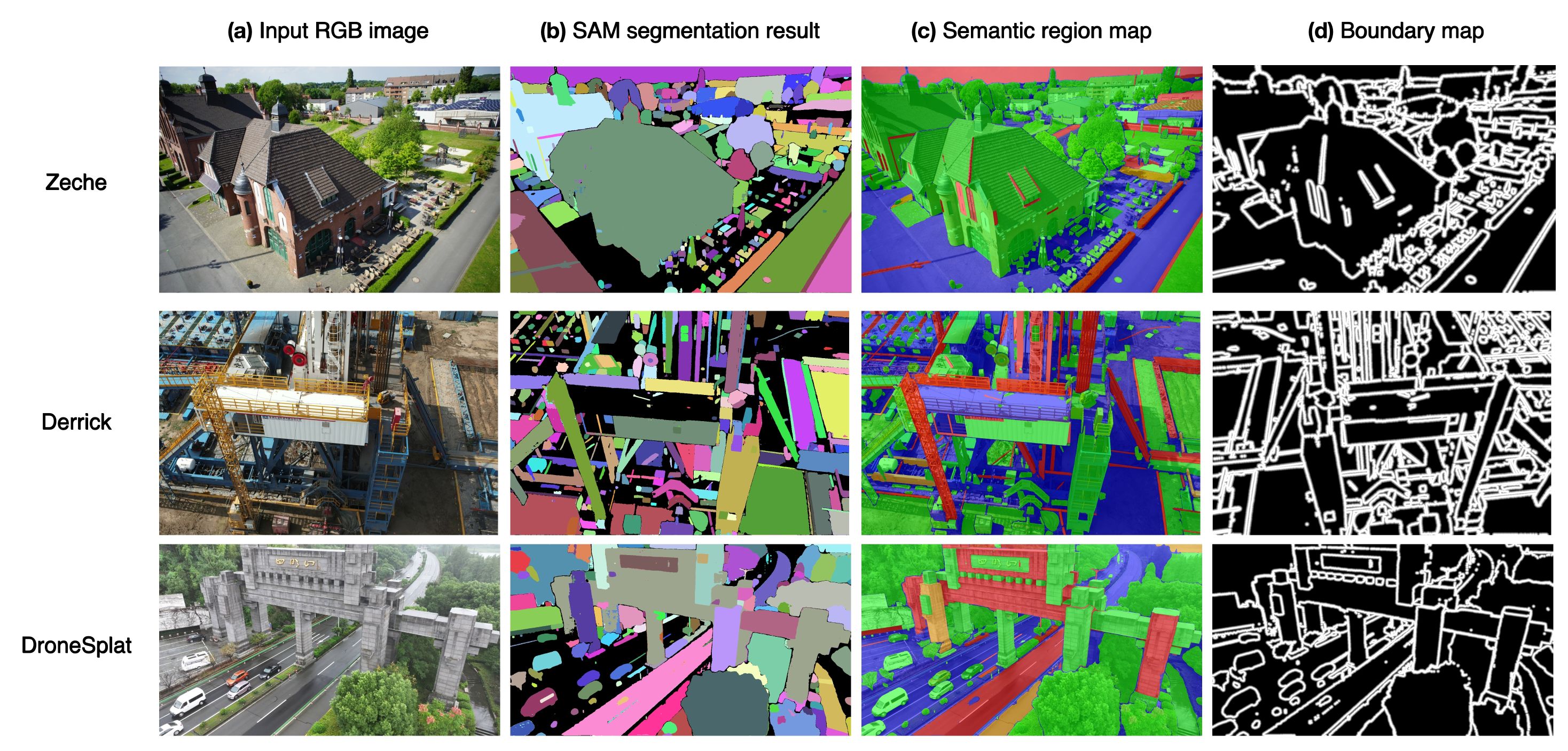}
\caption{Intermediate results of semantic preprocessing on the three datasets.}
\label{fig:preproc}
\end{figure}

Fig.~\ref{fig:preproc} shows the intermediate results of each stage of semantic preprocessing on the three datasets. SAM generates instance-level masks image by image (Fig.~\ref{fig:preproc}(b)). The masks are divided into four region types (Fig.~\ref{fig:preproc}(c)) by the classification rule of Section~\ref{sec:sgadc} (Eq.~(\ref{eq:classify})). The steel beams and cables of the derrick in the Derrick scene are stably recognized as linear structures (red) owing to their low compactness and large aspect ratio; high-gradient regions such as trees in the Zeche scene are assigned to the textured region (green); sky and road surfaces in the DroneSplat scene are assigned to the flat region (purple). The boundary map (Fig.~\ref{fig:preproc}(d)) forms continuous high-response bands at the junctions of adjacent masks, with the strongest responses on both sides of linear structures, which are the places where the original 3DGS most often produces ``broken lines'' and ``blurred edges''.

It should be noted that the goal of region classification is not pixel-accurate semantics. A small number of misclassifications (e.g., recognizing a thin shadow as a linear structure) only slightly changes the local threshold multiplier and does not alter the basic ``gradient-triggered'' logic of densification, so the influence on the final reconstruction is limited. Moreover, both the region map and the boundary map are generated offline.

\subsubsection{Adaptive density control}
To examine intuitively how SG-ADC redirects the Gaussian distribution, we directly compare the center positions, scales and shape parameters of the Gaussian models exported by 3DGS and SG-GS at the end of training (iteration 30,000) on the three datasets. For each scene, a unified orthographic top-view coordinate system is established by PCA of the 3DGS Gaussian centers, and both methods share the same transformation, cropping range, ROI, grid and color scale (Gaussians outside the cropping range or with non-finite coordinates are excluded from the statistics).

\begin{figure}[!t]
\centering
\includegraphics[width=\textwidth]{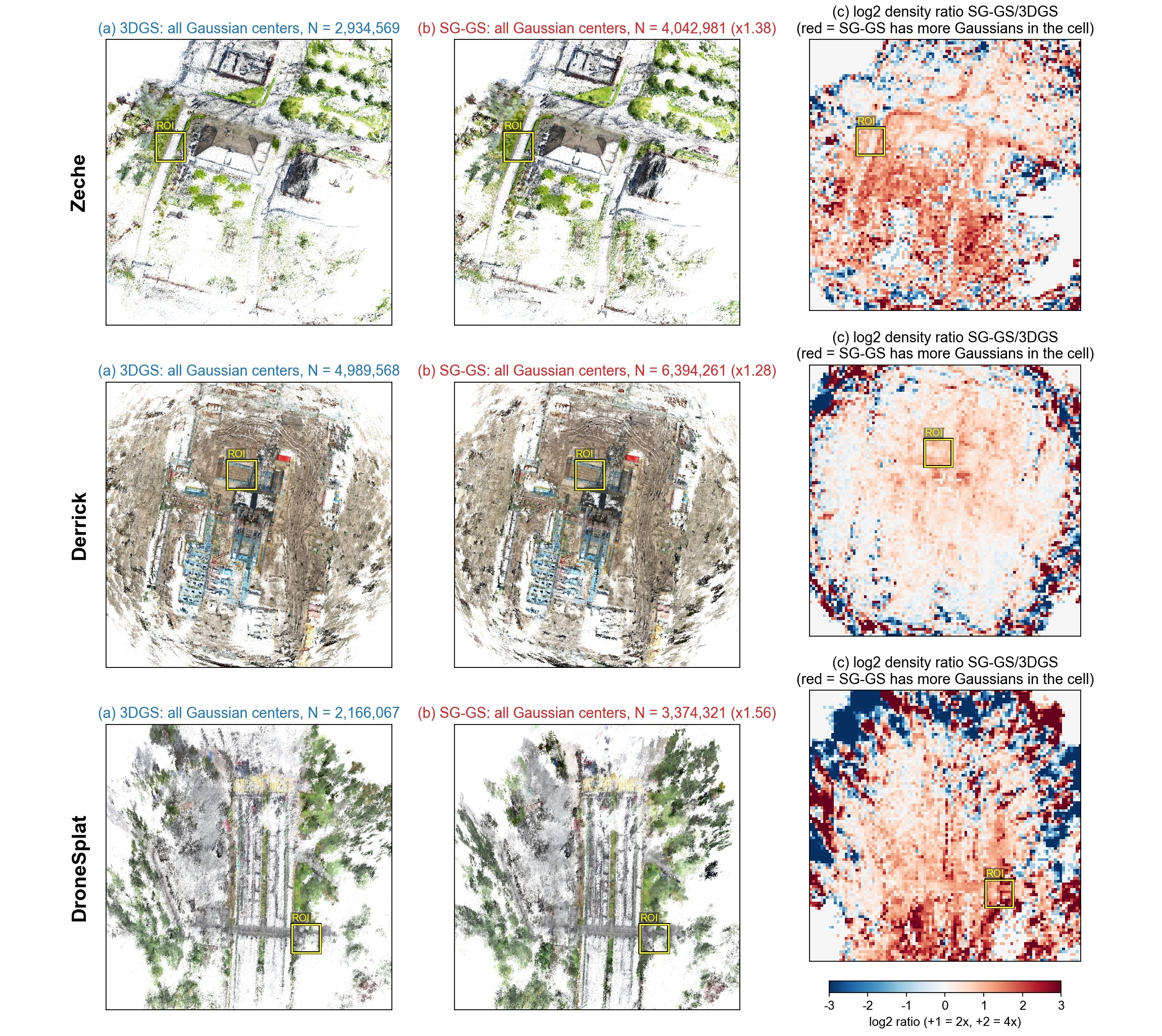}
\caption{The comparison of Gaussian distribution of 3DGS and SG-GS.}
\label{fig:panorama}
\end{figure}

Fig.~\ref{fig:panorama} gives the panoramic comparison of the three scenes: (a)(b) show all Gaussian centers of the two methods, (c) shows the logarithmic ratio of the number of SG-GS Gaussians to 3DGS Gaussians within the same grid cell (red means SG-GS has more Gaussians in the cell, blue means fewer), and the yellow boxes mark the selected ROIs. The Gaussian increment of SG-GS relative to 3DGS is clearly non-uniform: in Zeche and DroneSplat, high-gain regions are distributed in bands along structural directions such as building contours and road edges, while the ground is close to white or light blue; in Derrick, the gain concentrates on the central well-site facilities and the surrounding open ground is close to white, indicating that the new Gaussians are preferentially allocated to fine-structure regions. The outermost ring appears as red--blue interleaved noise because Gaussians are sparse there and it has no statistical significance.

\begin{figure}[!t]
\centering
\includegraphics[width=\textwidth]{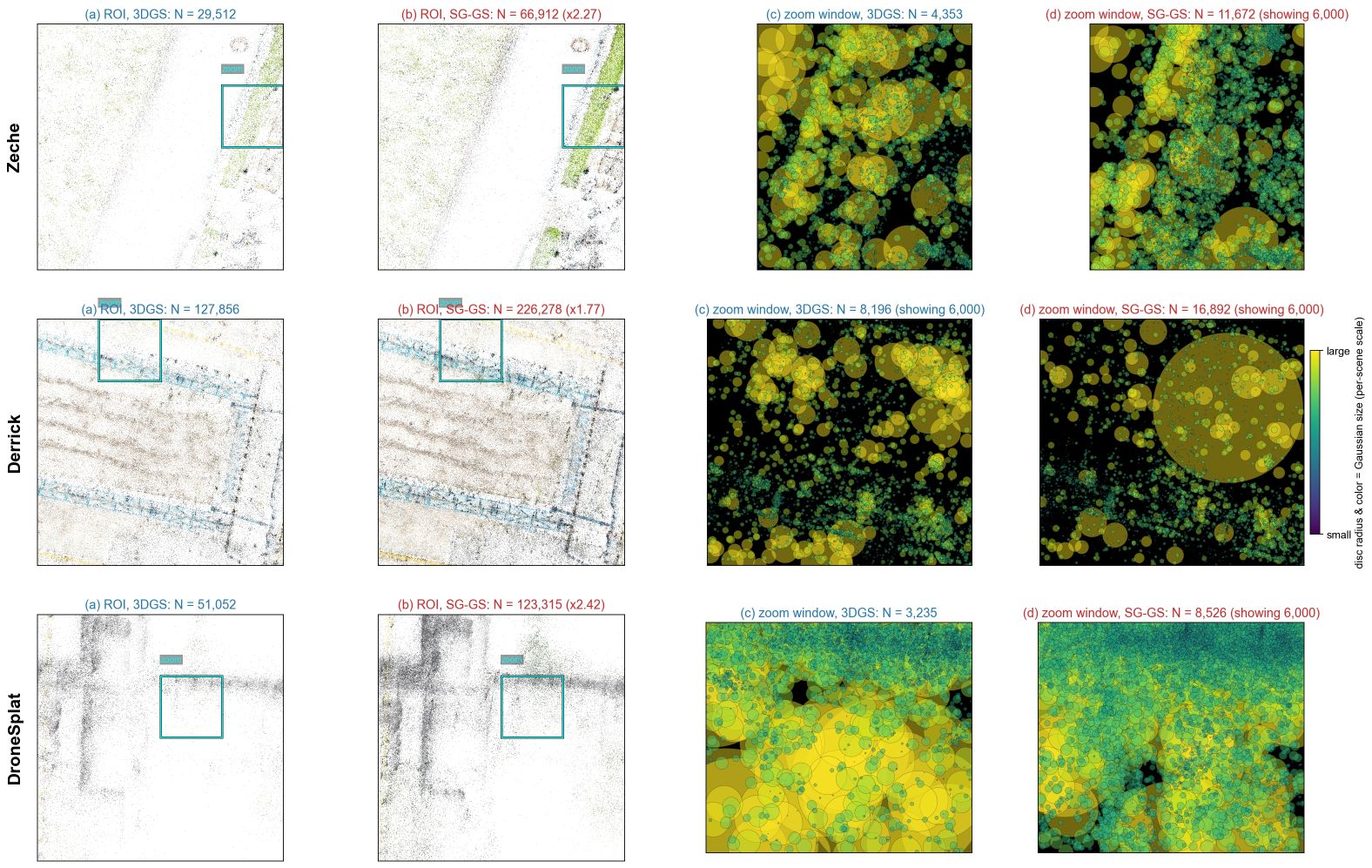}
\caption{Progressively zoomed comparison of the Gaussian distributions inside the ROIs.}
\label{fig:roi}
\end{figure}

Fig.~\ref{fig:roi} shows zoomed-in views of the ROIs. (a) and (b) present all Gaussian centers within the ROI, and (c) and (d) render each Gaussian inside the cyan zoom window as a disk. Both methods use the same window size and color scale.The newly added Gaussian centers in SG-GS follow the existing structural layout of 3DGS and fill the gaps between them, rather than appearing randomly in empty space. In the disk plots, baseline 3DGS represents structures in Zeche and DroneSplat with fewer Gaussians of highly uneven scales. In contrast, SG-GS samples the same structures more densely with more, smaller, and more uniformly sized Gaussians. For the Derrick scene, the disk sizes of the two methods are close, and the difference mainly lies in the number of Gaussians and their arrangement density.This is consistent with the mechanism described in Section~\ref{sec:sgadc}: the densification threshold multipliers for linear and boundary regions are lowered, so Gaussians in these regions split earlier and more frequently, covering the structures with finer Gaussians segment by segment.

\begin{figure}[!t]
\centering
\includegraphics[width=0.9\textwidth]{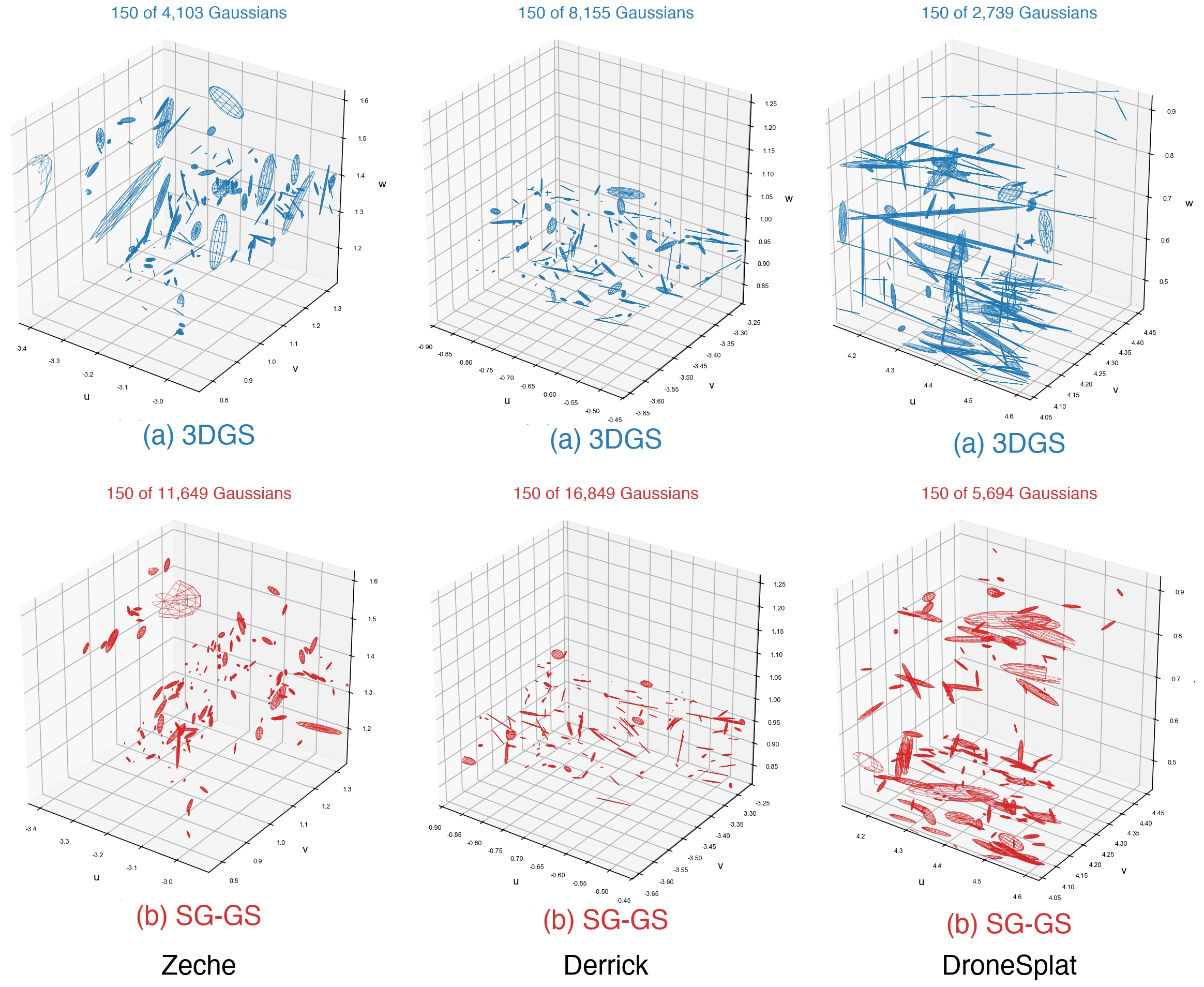}
\caption{3D shapes of the Gaussians inside the zoomed 3D cube window.}
\label{fig:3dshape}
\end{figure}

Fig.~\ref{fig:3dshape} further randomly samples 150 Gaussians inside the 3D cube corresponding to the zoom window and draws their true 3D shapes as ellipsoid wireframes, with the same viewpoint, range and height band for both methods. In Zeche and DroneSplat, the ellipsoids of 3DGS are markedly elongated and interpenetrating, which is the representation-level reason why the original 3DGS ``bridges'' linear structures with a few long Gaussians and thus produces ``blurred edges'' and ``broken lines'' in rendering; the ellipsoids of SG-GS are shorter and denser and are arranged segment by segment along the structures. In Derrick the shapes of the two are similar, and SG-GS mainly densifies the structures with more Gaussians of the same kind. It should be pointed out that the comparison in this section reflects the differences in number, scale and shape of the ``Gaussian representations'' of the two methods and is direct evidence of the SG-ADC mechanism, but it is not by itself equivalent to an improvement in rendering quality, which is given by the quantitative metrics and visual comparisons in Section~\ref{sec:quality}.

\subsection{Rendering quality evaluation}
\label{sec:quality}

\subsubsection{Quantitative evaluation}
The quantitative comparison on the three datasets is reported in Table~\ref{tab:comparison}. The compared methods include the original 3DGS \citep{kerbl20233dgs}, 2DGS \citep{huang20242dgs}, FastGS \citep{ren2025fastgs} and Scaffold-GS \citep{lu2024scaffoldgs}, all trained with the same data split and training configuration.

On the Derrick dataset, SG-GS achieves the best results on all three metrics: compared with 3DGS, PSNR increases by 1.34~dB (5.2\%), SSIM by 3.4\%, and LPIPS decreases by 21.4\%. 2DGS flattens the Gaussians into surfels in exchange for geometric accuracy, which limits its expressiveness in scenes with many thin lines, and all three of its metrics are below 3DGS; FastGS targets training efficiency and slightly improves PSNR, but its perceptual metric (LPIPS) deteriorates markedly, indicating that its acceleration sacrifices high-frequency details; Scaffold-GS obtains a considerable improvement through anchor-based structured Gaussians, but lacks a targeted design for linear structures, and its PSNR and LPIPS are still inferior to ours. The results on this dataset show that the semantic adaptive density control targeting linear structures yields the most direct benefit in such scenes.

\begin{table}[!t]
\centering
\caption{Quantitative comparison results.}
\label{tab:comparison}
\setlength{\tabcolsep}{3pt}
\resizebox{\textwidth}{!}{%
\begin{tabular}{lccccccccc}
\toprule
\multirow{2}{*}{\textbf{Method}} & \multicolumn{3}{c}{\textbf{Zeche}} & \multicolumn{3}{c}{\textbf{Derrick}} & \multicolumn{3}{c}{\textbf{DroneSplat}}\\
\cmidrule(lr){2-4}\cmidrule(lr){5-7}\cmidrule(lr){8-10}
 & PSNR\up & SSIM\up & LPIPS\dn & PSNR\up & SSIM\up & LPIPS\dn & PSNR\up & SSIM\up & LPIPS\dn\\
\midrule
3DGS        & 24.26 & 0.79 & 0.20 & 25.84 & 0.87 & 0.14 & 34.98 & 0.96 & 0.07\\
2DGS        & 24.36 & 0.76 & 0.24 & 25.56 & 0.85 & 0.18 & 31.55 & 0.94 & 0.11\\
FastGS      & 24.07 & 0.77 & 0.23 & 26.25 & 0.85 & 0.21 & 33.05 & 0.93 & 0.13\\
Scaffold-GS & \textbf{26.20} & \textbf{0.81} & 0.19 & 26.35 & 0.86 & 0.16 & 33.68 & 0.94 & 0.10\\
SG-GS (ours) & 24.93 & \textbf{0.81} & \textbf{0.18} & \textbf{27.18} & \textbf{0.90} & \textbf{0.11} & \textbf{36.51} & \textbf{0.98} & \textbf{0.05}\\
\bottomrule
\end{tabular}}
\end{table}

On the Zeche dataset, SG-GS improves PSNR by 0.67~dB (2.7\%) and SSIM by 2.5\% over 3DGS and reduces LPIPS by 10\%, with SSIM and LPIPS being the best among all methods. Notably, Scaffold-GS attains a higher PSNR on this dataset. Zeche is an aerial dataset with obvious illumination and exposure differences between views. The anchor features of Scaffold-GS are decoded into Gaussian attributes by MLPs per view, giving it view-adaptive appearance modeling that implicitly compensates exposure changes and thus an advantage in PSNR, which is based on pixel-wise mean squared error. Our method, like 3DGS, uses a static SH appearance model without per-view appearance compensation. However, on SSIM, which characterizes structural consistency, and LPIPS, which characterizes perceptual quality, our method is on par with or better than Scaffold-GS, indicating that its advantage lies in structure and detail rather than appearance fitting, consistent with the goal of structural refinement rather than appearance modeling. Moreover, Scaffold-GS introduces anchor features and several MLP decoders, increasing the model parameters and the decoding overhead at rendering time, whereas our method attaches no learnable parameters to the Gaussian primitives.

On the DroneSplat dataset, SG-GS again achieves the best results on all three metrics: compared with 3DGS, PSNR increases by 1.53~dB (4.4\%), SSIM by 2.1\%, and LPIPS decreases by 28.6\%. This dataset contains dynamic distractors such as vehicles, which pose an extra challenge to static scene reconstruction. Dynamic objects are inconsistent across views and induce persistent spurious gradients in flat background regions, based on which the original 3DGS generates many redundant Gaussians that try to fit the dynamic residues, forming floater artifacts. Our method raises the threshold of flat regions to 1.5 times, which significantly suppresses such erroneous densification triggered by dynamic inconsistency. This result further shows that the semantic-guided framework is robust to noise in the training data to a certain extent, extending the applicability of the method.

\subsubsection{Qualitative evaluation}
\begin{figure}[!t]
\centering
\includegraphics[width=0.9\textwidth]{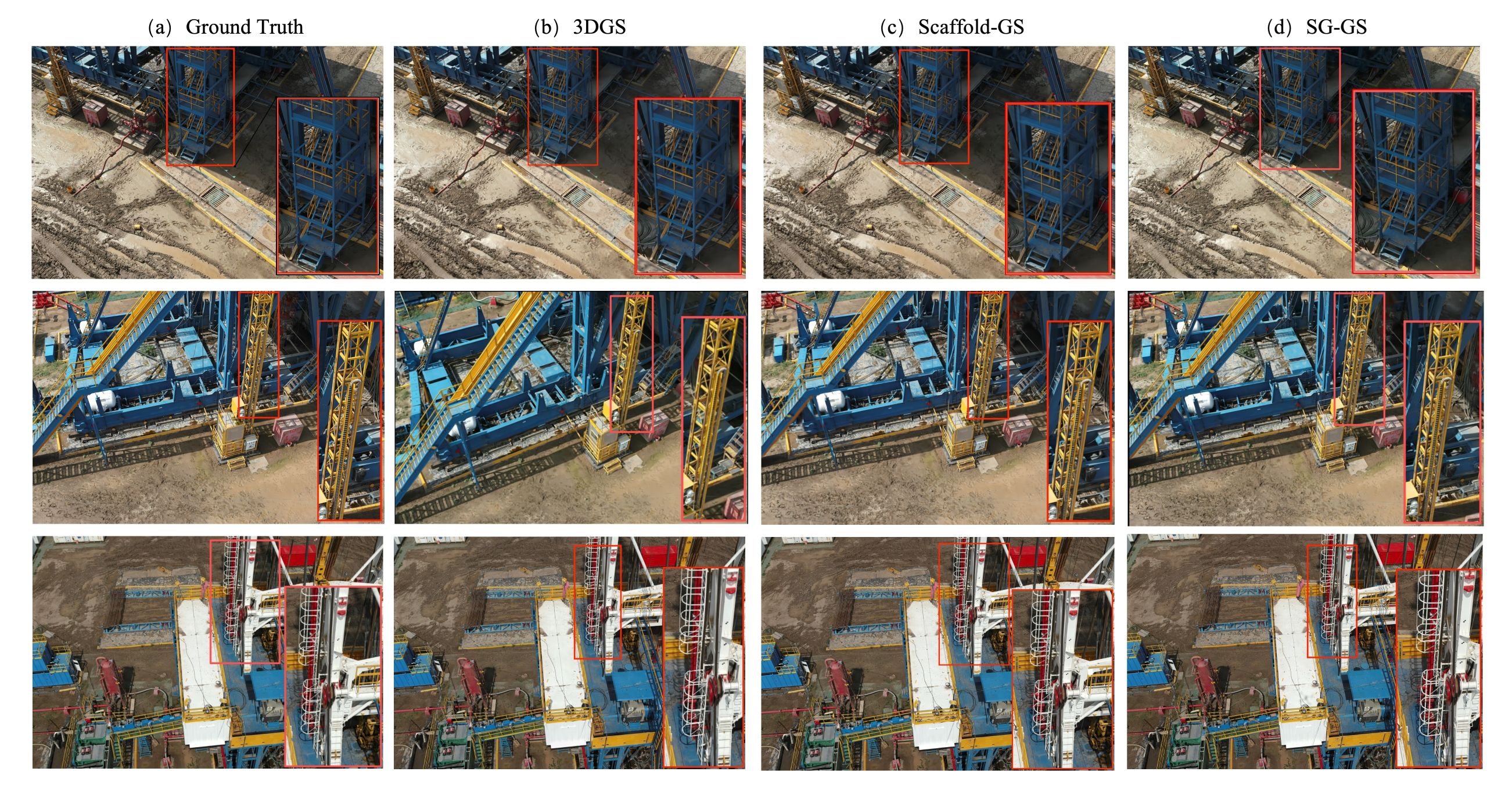}
\caption{Qualitative results on the Derrick dataset.}
\label{fig:qual_derrick}
\end{figure}

Fig.~\ref{fig:qual_derrick} shows the results on the Derrick dataset; from left to right are the ground truth, 3DGS, Scaffold-GS and SG-GS. The scene contains many linear structures such as transmission towers and pipelines. Such structures occupy few pixels and have weak rendering gradients, so they can hardly trigger densification under a globally uniform threshold; 3DGS can only approximate the thin geometry with a few large Gaussians, resulting in semi-transparent floaters and discontinuous structures. SG-GS lowers the gradient threshold of linear-structure regions to 0.5 times the base threshold and explicitly penalizes deviations of linear edges with the edge-preservation term (Eq.~(\ref{eq:ledge})), almost eliminating artifacts and blur and verifying the effectiveness of the semantic adaptive threshold mechanism.

\begin{figure}[!t]
\centering
\includegraphics[width=0.9\textwidth]{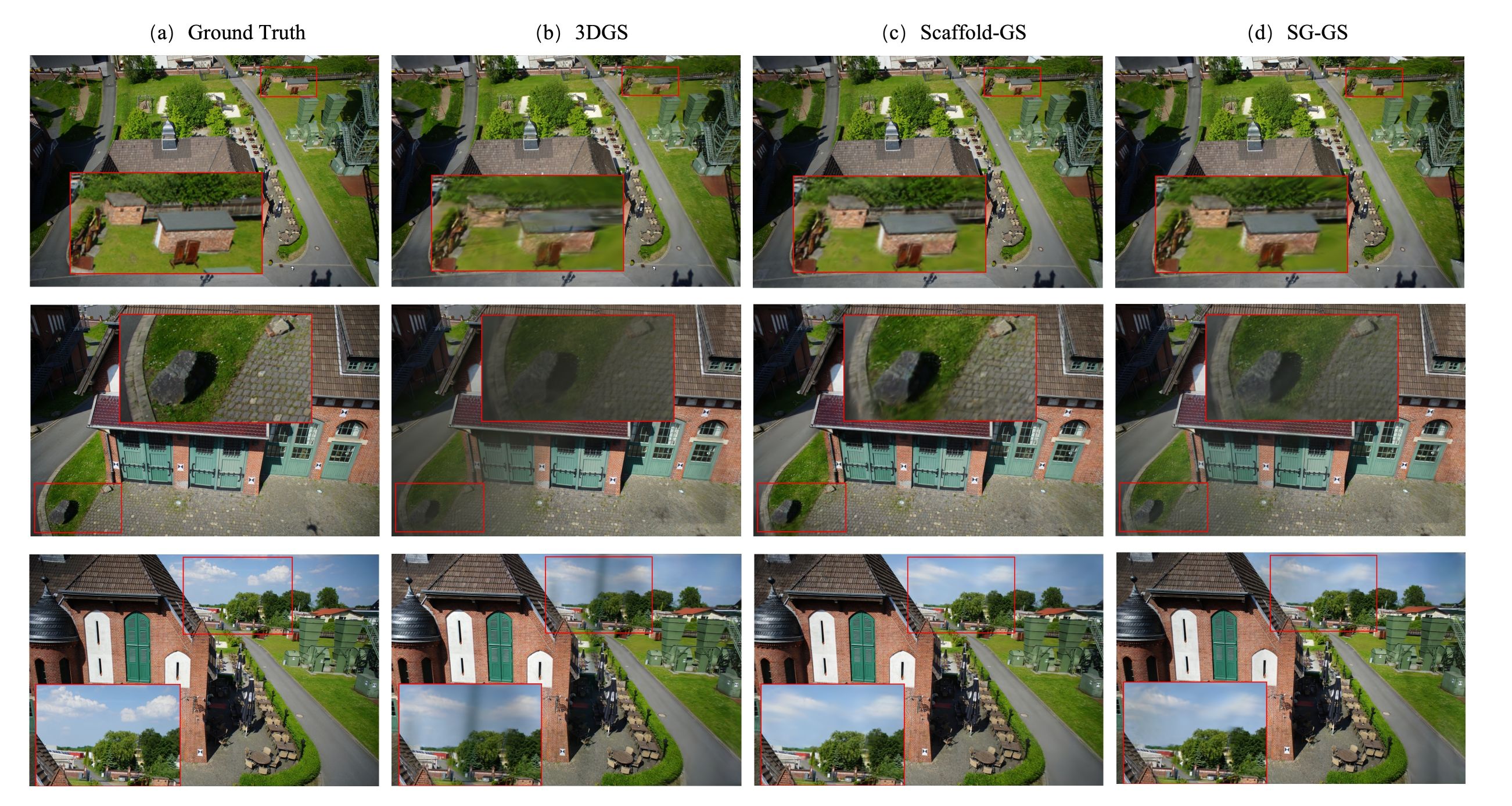}
\caption{Qualitative results on the Zeche dataset.}
\label{fig:qual_zeche}
\end{figure}

\begin{figure}[!t]
\centering
\includegraphics[width=0.9\textwidth]{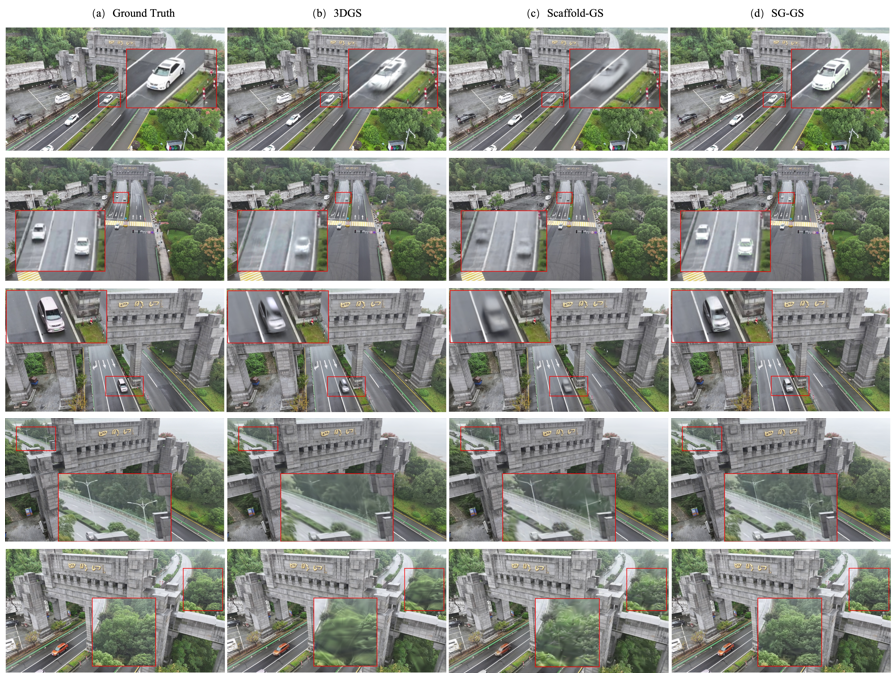}
\caption{Qualitative results on the DroneSplat dataset.}
\label{fig:qual_dronesplat}
\end{figure}

For the Zeche dataset, Fig.~\ref{fig:qual_zeche} shows the visual results. The scene has many object instances and complex textures; 3DGS is clearly blurred at the junction between the small house and the background and at object boundaries, because the uniform loss spreads the optimization capacity over large areas of low-information pixels and dilutes the error signal at the boundaries. SG-GS biases the optimization capacity towards key regions through the pixel importance weight map (Eq.~(\ref{eq:weightmap})) and supplements Gaussian primitives in the boundary neighborhood with the boundary-enhanced densification strategy, making both textures and building edges sharp.

Fig.~\ref{fig:qual_dronesplat} shows the rendering results on the DroneSplat dataset. The scene was captured by a drone and contains dynamic distractors such as vehicles, whose appearance and position are inconsistent between views; in the rendering they tend to appear as floating semi-transparent ghosts and local smearing, and the static textures near the residues are affected as well. Based on the semantic prior, SG-GS raises the gradient threshold of the other (flat) regions to 1.5 times the base threshold (Eq.~(\ref{eq:tau})), suppressing at the source the erroneous densification triggered by dynamic inconsistency: floater artifacts are markedly reduced and the road surface becomes clean and flat; meanwhile, the edge-preservation loss (Eq.~(\ref{eq:ledge})) keeps static structures such as building contours sharp and complete. The qualitative results corroborate the large lead in PSNR and LPIPS on this dataset in Table~\ref{tab:comparison}, indicating that the semantic-guided framework is robust to dynamic distractors.

\subsubsection{Efficiency analysis}
\begin{figure}[!t]
\centering
{\includegraphics[width=0.46\textwidth]{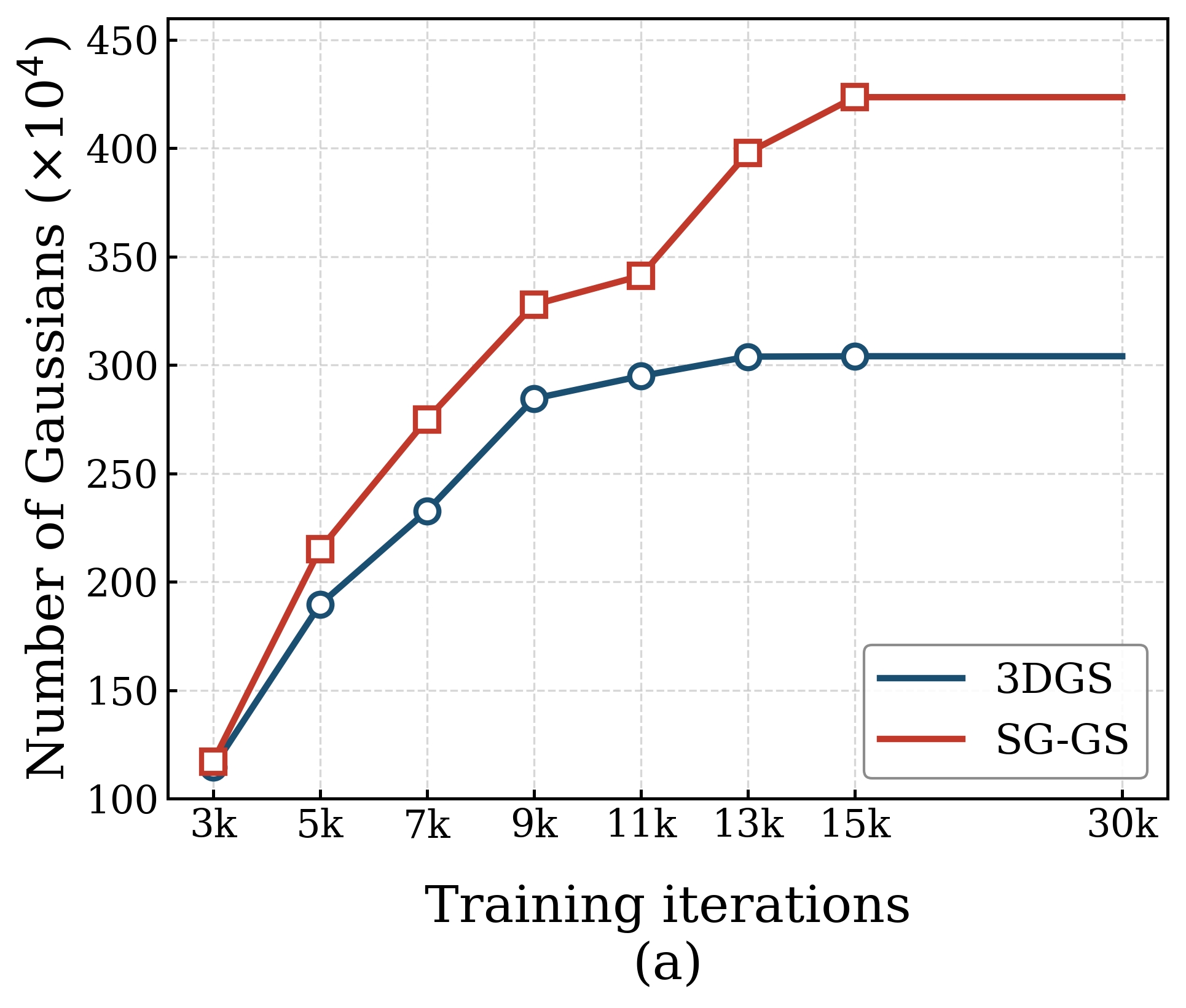}}\hfill
{\includegraphics[width=0.46\textwidth]{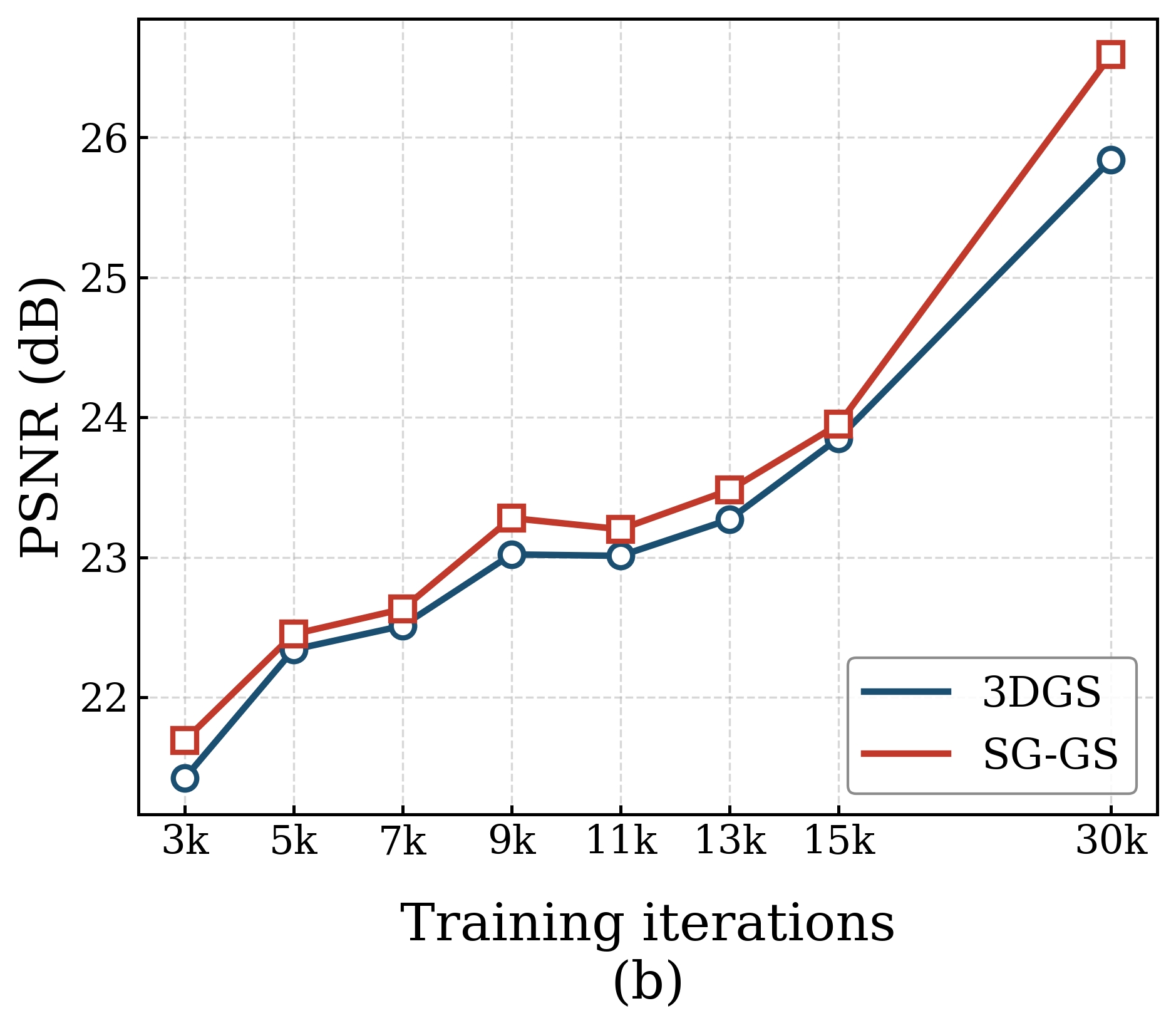}}
\caption{Evolution of the number of Gaussians and PSNR with the number of iterations.}
\label{fig:curves}
\end{figure}

Fig.~\ref{fig:curves} plots the evolution of the number of Gaussian primitives and the test PSNR with the number of iterations: (a) compares the number of Gaussians of 3DGS and SG-GS, and (b) compares their PSNR. The number of Gaussians of SG-GS is higher than that of 3DGS throughout training: the lowered thresholds in linear-structure and boundary regions make weak-gradient key regions trigger cloning and splitting earlier and more frequently. Importantly, these additional Gaussians are not indiscriminate redundant clones but are directed by the semantic prior to the regions where errors concentrate, such as linear structures and semantic boundaries. The PSNR curve shows that SG-GS leads 3DGS continuously once semantic guidance is enabled (iteration 3,000), and the lead widens further as the edge and frequency supervision ramp up, indicating that the additional Gaussians are effectively converted into reconstruction quality and that the quality gain per Gaussian is higher than the blind cloning of the baseline.

In terms of training time, under the same hardware and number of iterations, 3DGS takes 32.08~mins and SG-GS takes 44.9~mins, an increase of about 40\%. The extra cost comes from two sources: the increased total number of Gaussians raises the computation of rasterization and back-propagation in each iteration, and the edge convolution and FFT in MF-Loss introduce additional overhead. It should be noted that our method attaches no learnable features to the Gaussian primitives and introduces no decoding network at rendering time, which is fundamentally different from existing semantic fusion methods \citep{zhou2024feature3dgs,qin2024langsplat,ye2024gaussiangrouping} that continuously pay feature storage and decoding costs at the rendering stage. Thus, our method trades about 40\% more training time for stable accuracy improvements on the three datasets, a trade-off for fine-grained urban modeling applications where reconstruction quality is the primary goal.

Taking the quantitative and qualitative results on the three datasets together, the proposed method stably improves the reconstruction quality of urban scenes without attaching extra learnable features to the Gaussian primitives or introducing a decoding network, with especially significant improvements in linear structures and semantic boundary regions, which verifies the effectiveness and generalization ability of the semantic-guided framework.

\subsection{Ablation study}
\label{sec:ablation}

To evaluate the effectiveness of each module, ablation experiments are conducted on the Zeche dataset with unified parameter settings.

\subsubsection{Overall ablation}
SG-ADC and MF-Loss are added to the 3DGS baseline separately, and the results are listed in Table~\ref{tab:ablation}. Each module alone brings a stable gain. With SG-ADC only, PSNR increases by 0.49~dB over the baseline, SSIM by 0.02 and LPIPS decreases by 0.02, while the number of Gaussians grows from 3.0410 million to 4.1193 million, indicating that the module significantly strengthens the representation of structurally critical regions through semantic-directed densification at the cost of a certain growth in the number of Gaussians. With MF-Loss only, the number of Gaussians is essentially the same as the baseline while PSNR still increases by 0.34~dB, showing that semantic guidance on the loss side ensures sufficient and focused supervision of the primitives while barely changing the model capacity. When both are used, PSNR further increases to 24.93~dB, better than either module alone, and SSIM and LPIPS also remain the best, which shows that the mechanisms of the two modules are complementary and additive and verifies the rationality of jointly designing density control and loss supervision.

\begin{table}[!t]
\centering
\caption{Overall ablation results.}
\label{tab:ablation}
\small
\begin{tabular}{lcccc}
\toprule
\textbf{Model} & PSNR\up & SSIM\up & LPIPS\dn & \#Gaussians ($\times10^{4}$)\\
\midrule
3DGS (baseline) & 24.26 & 0.79 & 0.20 & 304.10\\
+SG-ADC & 24.75 & 0.81 & 0.18 & 411.93\\
+MF-Loss & 24.60 & 0.80 & 0.20 & 309.49\\
+SG-ADC+MF-Loss (full) & \textbf{24.93} & \textbf{0.81} & \textbf{0.18} & 423.60\\
\bottomrule
\end{tabular}
\end{table}

\subsubsection{Analysis of the semantic adaptive threshold}

Under the same training configuration, the globally uniform threshold and the region-adaptive threshold are compared, as listed in Table~\ref{tab:threshold}. Under the same densification-enhanced setting, the region-adaptive threshold improves PSNR by 0.35~dB and reduces LPIPS by 0.01 compared with the globally uniform threshold. Although uniformly lowering the global threshold also increases the number of Gaussians to 3.9253 million, the densification positions are decided solely by gradient magnitude, and a large part of the new budget is consumed in regions that are already well reconstructed, with limited improvement of weak-gradient linear structures. The region-adaptive threshold directs the densification budget to linear structures, semantic boundaries and texture-rich regions according to the semantic prior, while suppressing redundant generation in flat regions with a higher threshold, and achieves better reconstruction quality with only about 7.9\% more Gaussians.

\begin{table}[!t]
\centering
\caption{Ablation results of the adaptive threshold.}
\label{tab:threshold}
\small
\begin{tabular}{lcccc}
\toprule
\textbf{Threshold strategy} & PSNR\up & SSIM\up & LPIPS\dn & \#Gaussians ($\times10^{4}$)\\
\midrule
Globally uniform threshold & 24.58 & 0.81 & 0.19 & 392.53\\
Region-adaptive threshold & \textbf{24.93} & \textbf{0.81} & \textbf{0.18} & 423.60\\
\bottomrule
\end{tabular}
\end{table}

\subsubsection{Analysis of the multi-feature guided loss}

Taking the L1+D-SSIM baseline loss (with both SG-ADC and MF-Loss disabled, i.e., the 3DGS baseline) as the reference, the weight-map weighting, the edge-preservation loss and the frequency-domain high-frequency loss are introduced step by step with SG-ADC enabled; the results are listed in Table~\ref{tab:lossablation}. The weight-map weighting contributes the largest gain, improving PSNR by 0.52~dB over the baseline and verifying the redistribution of optimization capacity by the pixel importance weight map. Adding the edge-preservation loss on top slightly lowers PSNR, but LPIPS further decreases from 0.19 to 0.18, indicating that its benefit lies more in structure and perception. Further adding the frequency-domain high-frequency loss to form the complete MF-Loss raises PSNR back to 24.93~dB while keeping LPIPS at its best, indicating that the frequency constraint effectively recovers the lost high-frequency details. The three losses act on different levels and are ultimately complementary, jointly constituting the complete multi-feature guided supervision.

\begin{table}[!t]
\centering
\caption{Ablation results of the multi-feature guided loss.}
\label{tab:lossablation}
\begin{tabular}{lccc}
\toprule
\textbf{Loss function} & PSNR\up & SSIM\up & LPIPS\dn\\
\midrule
L1+D-SSIM (baseline loss) & 24.26 & 0.79 & 0.20\\
+Weight-map weighting & 24.78 & 0.81 & 0.19\\
+Edge-preservation loss & 24.58 & 0.81 & 0.18\\
+Frequency-domain loss (full MF-Loss) & \textbf{24.93} & \textbf{0.81} & \textbf{0.18}\\
\bottomrule
\end{tabular}
\end{table}

\subsubsection{Analysis of the phased progressive training}

\begin{figure}[!t]
\centering
\includegraphics[width=0.9\textwidth]{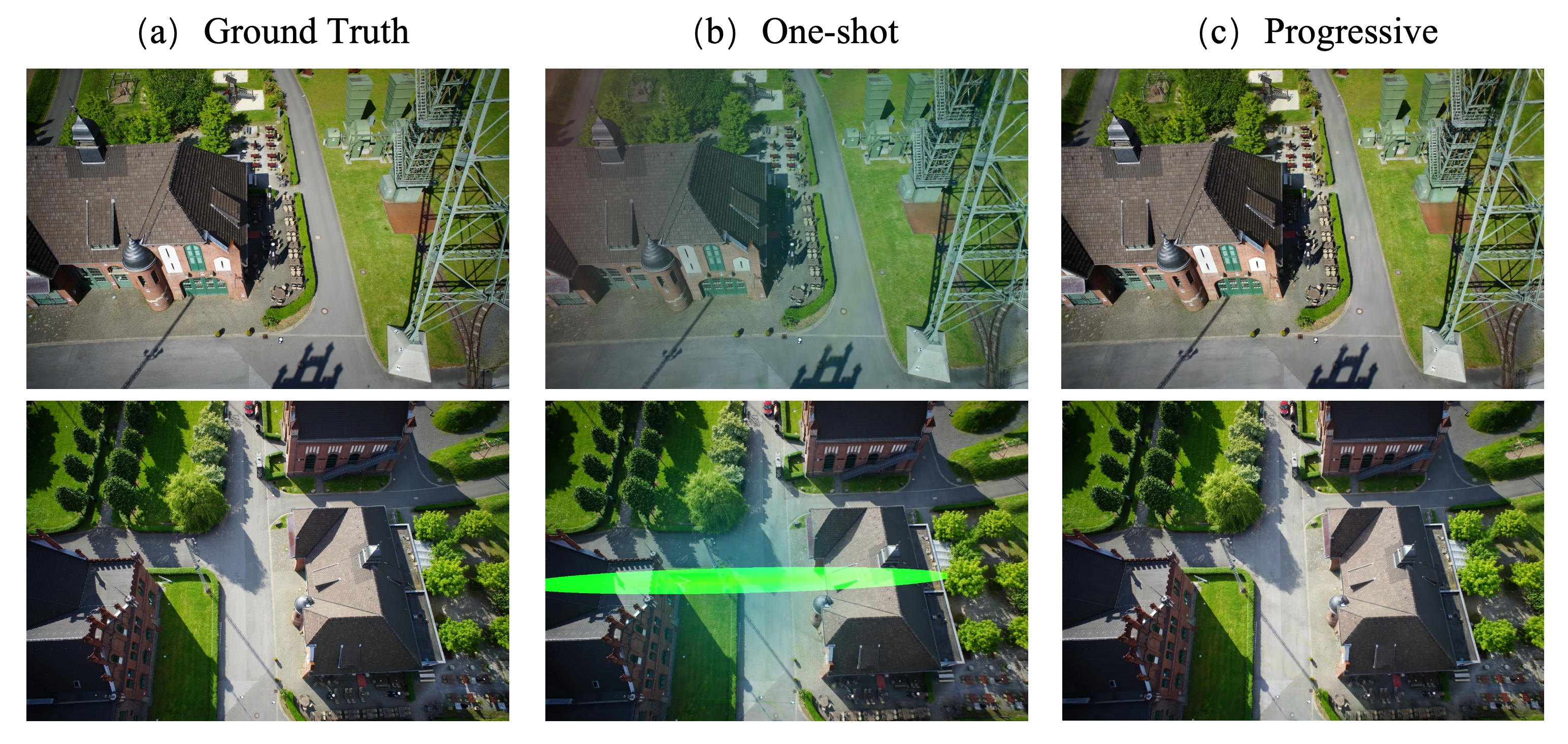}
\caption{Qualitative comparison of the training strategy ablation.}
\label{fig:ablschedule}
\end{figure}

Introducing all loss terms at full weight from the beginning of training (one-shot) is compared with the linear ramp of Eq.~(\ref{eq:schedule}) (progressive); the results are shown in Fig.~\ref{fig:ablschedule} and Table~\ref{tab:schedule}. The PSNR of the one-shot scheme is only 18.90~dB, far below the 24.93~dB of the progressive scheme and even below the 3DGS baseline, and LPIPS deteriorates sharply as well. The reason is that the rendering is noisy in early training, so the gradients of the edge and frequency terms are dominated by rendering noise and produce many erroneous supervision signals; these erroneous gradients are further amplified by the densification mechanism, driving the Gaussian distribution away from the true geometry and eventually destabilizing convergence. The progressive strategy releases fine supervision signals in the order of skeleton first and details later, introducing the edge and frequency constraints only after the geometric skeleton has stabilized, which ensures that each supervision term takes effect at the right time.

\begin{table}[!t]
\centering
\caption{Ablation results of the training strategy.}
\label{tab:schedule}
\begin{tabular}{lccc}
\toprule
\textbf{Training strategy} & PSNR\up & SSIM\up & LPIPS\dn\\
\midrule
One-shot introduction & 18.90 & 0.74 & 0.29\\
Progressive introduction & \textbf{24.93} & \textbf{0.81} & \textbf{0.18}\\
\bottomrule
\end{tabular}
\end{table}

\section{Conclusion}
\label{sec:conclusion}

To address the coarse density control strategy and semantically blind loss function of the original 3DGS in reconstructing complex urban structures, this paper proposes SG-GS, a semantic-guided fine-grained 3DGS reconstruction method that jointly optimizes density control and loss supervision. A semantic-constrained adaptive density control mechanism is built, which allocates gradient thresholds according to region type and adds a boundary-enhanced densification strategy, so that new Gaussians concentrate on key regions such as linear structures and semantic boundaries. A multi-feature guided loss function is designed, which weights the reconstruction loss per pixel with a pixel importance weight map, adds edge-preservation and frequency-domain high-frequency constraints, and ensures convergence stability through a phased progressive training strategy. Experiments on three urban scene datasets show that the proposed method stably outperforms the original 3DGS and effectively mitigates artifacts on linear structures and boundary blurring. The method places semantic guidance entirely on the training side, attaches no learnable parameters to the Gaussian primitives, introduces no decoding overhead at rendering time.

The method also has limitations. First, appearance modeling still relies on static spherical harmonics, so its PSNR on datasets with obvious illumination and exposure changes is inferior to Scaffold-GS, which has view-adaptive appearance capability. Second, region classification depends on the segmentation quality of SAM; in weakly textured or strongly reflective regions where segmentation fails, the guidance signal may degrade to the baseline behavior. Future work will explore the joint optimization of semantic guidance with view-adaptive appearance modeling and surface reconstruction, as well as block-wise training strategies for large-scale scenes.

\section*{Acknowledgements}
This research was funded by the National Natural Science Foundation of China (No. 42671574, 42371442, 42571483); the Shenzhen Science and Technology Program (No. JCYJ20250604181614019); and Guangdong Basic and Applied Basic Research Foundation (No. 2025A1515011725).


\bibliographystyle{elsarticle-harv}
\bibliography{refs}

\end{document}